\documentclass[twoside,11pt]{article}

\usepackage{jmlr2e}

\usepackage{color}
\usepackage{url}

\jmlrheading{x}{2012}{xxx}{xx/xx}{xx/xx}{H.C. Burger, C.J. Schuler and S. Harmeling}

\ShortHeadings{Image denoising with multi-layer perceptrons, part 1}{H.C. Burger, C.J. Schuler and S. Harmeling}
\firstpageno{1}

\graphicspath{
  {figures/}
}

\begin{document}

\title{Image denoising with multi-layer perceptrons, part 1:\\
comparison with existing algorithms and with bounds}

\author{\name Harold Christopher Burger \email burger@tuebingen.mpg.de \\
       \name Christian J. Schuler \email cschuler@tuebingen.mpg.de \\
       \name Stefan Harmeling  \email harmeling@tuebingen.mpg.de \\
       \addr Max Planck Institute for Intelligent Systems\\
       Spemannstr. 38\\
       72076 T\"{u}bingen, Germany
       }

\editor{}

\maketitle

\begin{abstract}
Image denoising can be described as the problem of mapping from a noisy image
to a noise-free image. The best currently available denoising methods
approximate this mapping with cleverly engineered algorithms. In this work we
attempt to learn this mapping directly with plain multi layer perceptrons (MLP)
applied to image patches. We will show that by training on large image
databases we are able to outperform the current state-of-the-art image
denoising methods. In addition, our method achieves results that are superior
to one type of theoretical bound and goes a large way toward closing the gap
with a second type of theoretical bound.  Our approach is easily adapted to
less extensively studied types of noise, such as mixed Poisson-Gaussian noise,
JPEG artifacts, salt-and-pepper noise and noise resembling stripes,  for which
we achieve excellent results as well. We will show that combining a
block-matching procedure with MLPs can further improve the results on certain
images. In a second paper \citep{burgerjmlr2}, we detail the training trade-offs
and the inner mechanisms of our MLPs.
\end{abstract}

\tableofcontents
\section{Introduction}
\label{sec:intro}
\begin{figure}[ht]
  \centering
    \begin{tabular}{cc}
      \includegraphics[width=0.48\textwidth]{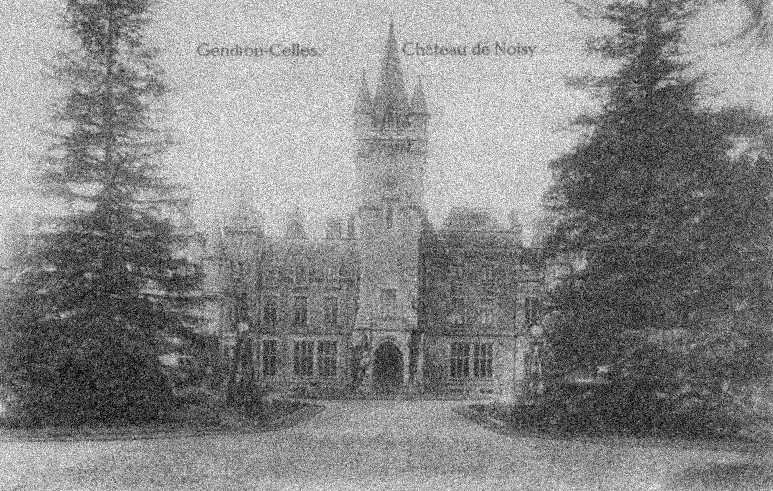} &
      \includegraphics[width=0.48\textwidth]{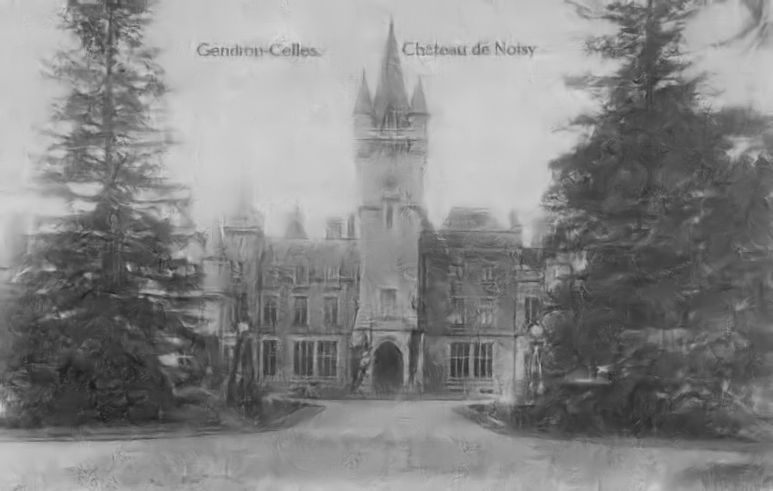} \\
      Noisy castle & Clean castle 
    \end{tabular}
  \caption{The goal of image denoising is to find a clean version of the noisy
  input image.}
  \label{fig:noisy}
\end{figure}
Images are invariably corrupted by some degree of noise. The strength and type of
noise corrupting the image depends on the imaging process. In scientific
imaging, one sometimes needs to take images in a low photon-count setting, in
which case the images are corrupted by mixed Poisson-Gaussian
noise~\citep{luisier2011image}. Magnetic resonance images are usually corrupted
by noise distributed according to the Rice
disitribution~\citep{gudbjartsson1995rician}. For natural images captured by a
digital camera, the noise is usually assumed to be additive, white and
Gaussian-distributed (AWG noise), see for example~\citet{elad2006image,dabov2007image}.

An image denoising procedure takes a noisy image as input and estimates an
image where the noise has been reduced.  Numerous and diverse approaches exist:
Some  selectively smooth parts of a noisy
image~\citep{tomasi2002bilateral,weickert1998anisotropic}.  Other methods rely
on the careful shrinkage of wavelet
coefficients~\citep{simoncelli1996noise,portilla2003image}. A 
conceptually similar approach is to denoise image patches by trying to
approximate noisy patches using a sparse linear combination of elements of a
learned dictionary~\citep{aharon2006rm,elad2006image}.
BM3D~\citep{dabov2007image} is a very successful approach to denoising and is
often considered state-of-the art. The approach does not rely on a
probabilistic image prior but rather exploits the fact that images are often
self-similar: A given patch in an image is likely to be found elsewhere in the
same image. In BM3D, several similar-looking patches of a noisy image are
denoised simultaneously and \emph{collaboratively}: Each noisy patch helps to
denoise the other noisy patches. The algorithm does not rely on learning from a
large dataset of natural images; excellent denoising results are achieved
through the design of the algorithm. While BM3D is a well-engineered algorithm,
could we also \emph{automatically learn} an image denoising procedure purely
from training examples consisting of pairs of noisy and noise-free patches?

%%%%%%%%%%%%%%%%%%%%%%%%%%%%%%%%%%%%%%%%%%%%%%%%%%%%%%%%%%%%%%%
\paragraph{Denoising as a function:} In image denoising, one is given a noisy
version of a clean image, where the noise is for instance i.i.d. Gaussian
distributed with known variance (AWG noise).  The goal is to find the clean
image, given only the noisy image. We think of denoising as a function that
maps a noisy image to a cleaner version of that image. However, the
complexity of a mapping from images to images is large, so in practice we chop the image into
possibly overlapping \emph{patches} and learn a mapping from a noisy patch to a clean patch.  To denoise
a given image, all image patches are denoised separately by that map.  The denoised image
patches are then combined into a denoised image.

The size of the patches affects the quality of the denoising function.
If the patches
are small and the noise level is high, many clean patches are a potential
explanation for a given noisy patch. In other words, adding noise to a clean
patch is not injective and therefore also not invertible. It is therefore
almost impossible to find a perfect denoising function. Lowering the noise and
increasing the size of the patches alleviates this problem: Fewer clean patches
are a potential explanation for a given noisy image~\citep{levin2010natural}.
At least in theory, better denoising results are therefore achievable with large patches than with
small patches.

In practice, the mapping from noisy to clean patches cannot
be expressed using a simple formula. However, one can easily generate
\emph{samples}: Adding noise to a patch creates an argument-value pair, where
the noisy patch is the argument of the function and the noise-free patch is the
value of the function.

The aim of this paper is to \emph{learn} the denoising function. For this, we
require a \emph{model}. The choice of the model is influenced by the function
to approximate. Complicated functions require models with high \emph{capacity},
whereas simple functions can be approximated using a model with low capacity.
The dimensionality of the problem, which is defined by the size of the patches,
is one measure of the difficulty of approximation. One should therefore expect
that models with more capacity are required when large image patches are used.
A higher dimensionality also usually implies that more \emph{training data} is
required to learn the model, unless the problem is intrinsically of low
dimension.

We see that a trade-off is necessary: Very small patches lead to a function that
is easily modeled, but to bad denoising results. Very large patches potentially
lead to better denoising results, but the function might be difficult to
model.  

This paper will show that it is indeed
possible to achieve state-of-the-art denoising performance with a plain multi
layer perceptron (MLP) that maps noisy patches onto noise-free ones. This is
possible because the following factors are combined:
\begin{itemize}
\item The capacity of the MLP is chosen large enough, meaning that it consists
of enough hidden layers with sufficiently many hidden units.
\item The patch size is chosen large enough, so that a patch contains enough
  information to recover a noise-free version. This is in agreement
  with previous findings~\citep{levin2010natural}.
\item The chosen training set is large enough. Training examples are
  generated on
  the fly by corrupting noise-free patches with noise.
\end{itemize}
Training high capacity MLPs with large training sets is feasible using
modern Graphics Processing Units (GPUs). \citet{burgerjmlr2} contains a
detailed analysis of the trade-offs during training.
\medskip

\paragraph{Contributions:} We present a patch-based denoising algorithm
that is \emph{learned} on a large dataset with a plain neural network.
Additional contributions of this paper are the following.
\begin{enumerate}
  \item We show that the state-of-the-art is improved on AWG noise. This is
  done using a  thorough evaluation on $2500$ test images,
  \item excellent results are obtained on mixed Poisson-Gaussian noise, JPEG
  artifacts, salt-and-pepper noise and noise resembling stripes, and
  \item We present a novel ``block-matching'' multi-layer perceptron and
  discuss its strengths and weaknesses.
  \item We relate our results to recent theoretical work on the limits of
  denoising~\citep{chatterjee2010denoising, levin2010natural, levin2012patch}.
  We will show that two of the bounds described in these papers cannot be
  regarded as hard limits. We make important steps towards reaching the third
  proposed bound.
\end{enumerate}
While we have previously shown that MLPs can achieve outstanding image
denoising results \citep{burger2012image}, in this work we present
significantly improved results compared to our previous work as well as more
thorough experiments.

\section{Related work}
The problem of removing noise from natural images has been extensively studied,
so methods to denoise natural images are numerous and diverse.
\citet{estrada2009stochastic} classify denoising algorithms into three
categories: 
\begin{enumerate}
  \item The first class of algorithms rely on smoothing parts of the noisy
  image~\citep{rudin1992nonlinear,weickert1998anisotropic,tomasi2002bilateral}
  with the aim of ``smoothing out'' the noise while preserving image details.
  \item The second class of algorithms exploits the fact that different patches
  in the same image are often similar in
  appearance~\citep{dabov2007image,buades2006review}. 
  \item The third class of denoising algorithms exploit learned image
  statistics. A natural image model is typically learned on a noise-free
  training set (such as the Berkeley segmentation dataset) and then exploited
  to denoise images~\citep{roth2009fields,weiss2007makes,jain2009natural}.  In
  some cases, denoising might involve the careful shrinkage of coefficients.
  For
  example~\citet{simoncelli1996noise,chang2002adaptive,pizurica2002joint,portilla2003image}
  involve shrinkage of wavelet coefficients. Other methods denoise small images
  patches by representing them as sparse linear combinations of elements of a
  learned dictionary \citep{elad2006image,mairal2008sparse,mairal2010non}. 
\end{enumerate}

\paragraph{Neural networks:} Neural networks belong to the category relying on
learned image statistics. They have already been used to denoise
images~\citep{jain2009natural} and belong in the category of learning-based
approaches.  The networks commonly used are of a special type, known as
\emph{convolutional neural networks} (CNNs)~\citep{lecun1998gradient}, which
have been shown to be effective for various tasks such as hand-written digit
and traffic sign recognition~\citep{sermanettraffic}.  CNNs exhibit a structure
(local receptive fields) specifically designed for image data.  This allows for
a reduction of the number of parameters compared to plain multi layer
perceptrons while still providing good results. This is useful when the amount
of training data is small. On the other hand, multi layer perceptrons are
potentially more powerful than CNNs: MLPs can be thought of as universal
function
approximators~\citep{cybenko1989approximation,hornik1989multilayer,funahashi1989approximate,leshno1993multilayer},
whereas CNNs restrict the class of possible learned functions. 

A different kind of neural network with a special architecture (containing
a \emph{sparsifying logistic}) is used in~\citep{marc2007unified} to denoise
image patches. A small training set is used. Results are reported for strong
levels of noise. 
It has also been attempted to denoise images by applying multi layer perceptrons on 
wavelet coefficients~\citep{zhang2005image}. The use of wavelet bases can be seen as an 
attempt to incorporate prior knowledge about images.

Denoising auto-encoders~\citep{vincent2010stacked} also use the idea of using
neural networks for denoising. Denoising auto-encoders are a special type of
neural network which can be trained in an unsupervised fashion. Interesting
features are learned by the units in the hidden layers. For this, one exploits
the fact that training pairs can be generated cheaply, by somehow corrupting
(such as by adding noise to) the input.  However, the goal of these networks is not
to achieve state-of-the-art results in terms of denoising performance, but
rather to learn representations of data that are useful for other tasks.
Another difference is that typically, the noise used is not AWG noise, but
salt-and-pepper noise or similar forms of noise which ``occlude'' part of the
input. Denoising auto-encoders are learned layer-wise and then \emph{stacked},
which has become the standard approach to deep learning~\citep{hinton2006fast}.
The noise is applied on the output of the previously learned layer. This is
different from our approach, in which the noise is always applied on the input
patch only and all layers are learned \emph{simultaneously}.

Our approach is reminiscent of deep learning approaches because we also employ
several hidden layers. However, the goal of deep learning is to learn several
levels of representations, corresponding to a hiearchy of features,
see~\citet{bengio2009learningdeep} for an overview. In this work we are mainly
interested in image denoising results.

\medskip 

\paragraph{Innovations in this work:} Most methods based on neural
networks make assumptions about natural images. Instead, we show that
state-of-the-art results can be obtained by imposing no such assumptions, but
by relying on a pure \emph{learning} approach.

\section{Learning to denoise}
In Section~\ref{sec:intro}, we defined the denoising problem as learning the
mapping from a noisy patch to a cleaner patch. For this, we require a model. In
principle, different models could be used, but we will use MLPs for that
purpose. We chose MLPs over other models because of their ability to handle
large datasets.

\subsection{Multi layer perceptrons (MLPs)}
\begin{figure}[htbp]
  \centering
  \includegraphics[width=0.8\columnwidth]{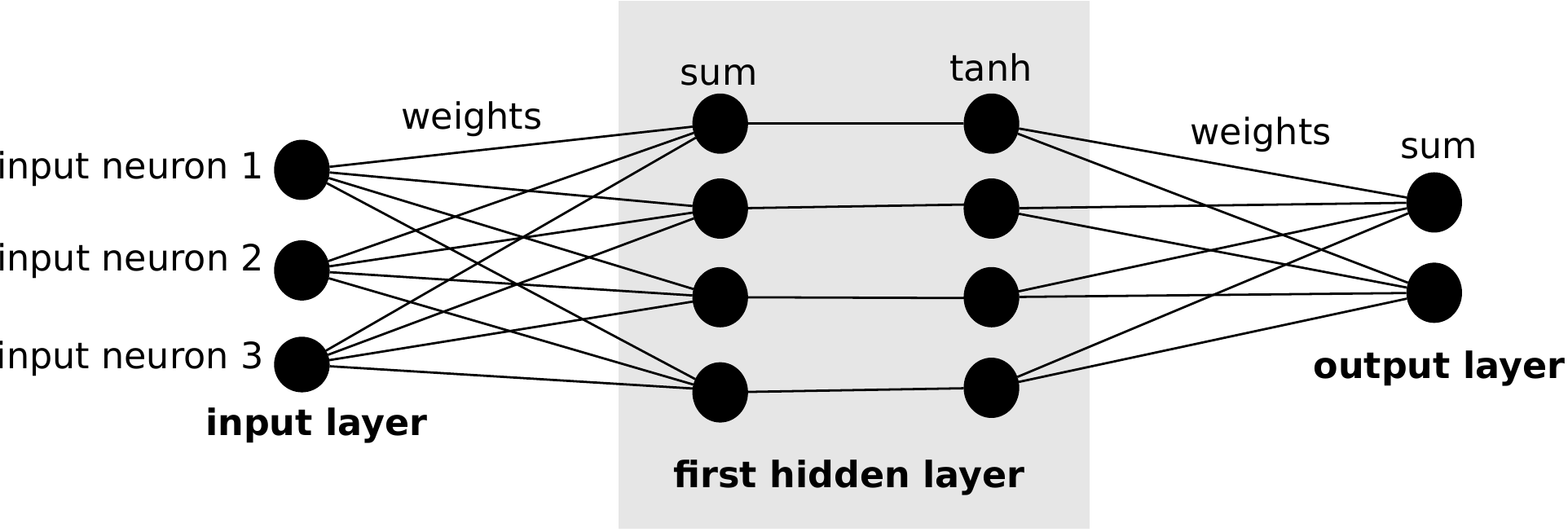} 
  \caption{A graphical representation of a (3,4,2)-MLP.}
  \label{fig:mlp}
\end{figure}
A \emph{multi layer perceptron} (MLP) is a nonlinear function that
maps vector-valued input via several hidden layers to vector-valued output.
For instance, an MLP with two hidden layers can be written as,
\begin{equation}
\label{eqn:mlp}
  f(x) = b_3 + W_3\tanh(b_2 + W_2\tanh(b_1 + W_1 x)).
\end{equation}
The weight matrices $W_1, W_2, W_3$ and vector-valued biases $b_1, b_2, b_3$
parameterize the MLP, the function $\tanh$ operates component-wise. The
\emph{architecture} of an MLP is defined by the number of hidden layers and by
the layer sizes.  For instance, a (256,2000,1000,10)-MLP has two hidden layers.
The input layer is 256-dimensional, i.e.~$x\in\Re^{256}$.  The vector
$v_1=\tanh(b_1+W_1x)$ of the first hidden layer is 2000-dimensional, the vector
$v_2=\tanh(b_2 + W_2v_1)$ of the second hidden layer is 1000-dimensional, and
the vector $f(x)$ of the output layer is 10-dimensional.  Commonly, an MLP is
also called \emph{feed-forward neural network}. MLPs can also be represented
graphically, see Figure~\ref{fig:mlp}. All our MLPs are \emph{fully connected},
meaning that the weight matrices $W_{i}$ are dense. One could also imagine MLPs
which are not fully connected, using sparse weight matrices. Sparsely connected
MLPs have the advantage of being potentially computationally easier to train and 
evaluate.

MLPs belong to the class of \emph{parametric} models, the parameters being
estimated during learning. However, the number of parameters in MLPs is often
so large that they are extremely flexible.

\subsection{Training MLPs for image denoising}
To train an MLP that maps noisy image patches onto clean image patches where
the noise is reduced or even removed, we estimate the parameters by training on
pairs of noisy and clean image patches using \emph{stochastic gradient descent}
\citep{lecun1998efficient}.  

More precisely, we randomly pick a clean patch $x$ from an image dataset and
generate a corresponding noisy patch $y$ by corrupting $x$ with noise, for
instance with additive white Gaussian (AWG) noise. We then feed the noisy patch
$x$ into the MLP to compute $f(x)$, representing an estimate of the clean patch
$x$. The MLP parameters are then updated by the \emph{backpropagation}
algorithm \citep{rumelhart1986learning} minimizing the squared error between the
mapped noisy patch $f(x)$ and the clean patch $y$, i.e.~minimizing pixel-wise
$(f(x)-y)^2$. We choose to minimize 
the mean squared error since it is monotonically related to the PSNR, which is 
the most commonly used measure of image quality. Thus
minimizing the squared error will maximize PSNR values.

To make backpropagation efficient, we apply various common neural
network tricks
\citep{lecun1998efficient}:
\begin{enumerate}
  \item Data normalization: The pixel values are transformed to have
  approximately mean zero and variance close to one.  More precisely, assuming
  pixel values between $0$ and $1$, we subtract $0.5$ and multiply by $0.2$.
  \item Weight initialization: We use the ``normalized initialization''
  described by~\citet{GlorotAISTATS2010}. The weights are sampled from a uniform
  distribution: \begin{equation} w \sim U\left[ -\frac{\sqrt{6}}{\sqrt{n_{j} +
  n_{j+1}}}, \frac{\sqrt{6}}{\sqrt{n_{j} + n_{j+1}}} \right], \end{equation}
  where $n_{j}$ and $n_{j+1}$ are the number of neurons in the input side and
  output side of the layer, respectively. Combined with the first trick, this
  ensures that both the linear and the non-linear parts of the sigmoid function
  are reached.
  \item Learning rate division: In each layer, we divide the learning rate by
  $N$, the number of input units of that layer. This allows us to change the
  number of hidden units without modifying the learning rate.
\end{enumerate}
The basic learning rate was set to $0.1$ for most experiments. The training
procedure is discussed in more detail in~\citet{burgerjmlr2}.

\subsection{Number of hidden layers}
The number hidden layers as well as the number of neurons per hidden layer
control the capacity of the model. No more than a single hidden layer is needed
to approximate any function, provided that layer contains a sufficient number
of neurons ~\citep{cybenko1989approximation, hornik1989multilayer,
funahashi1989approximate, leshno1993multilayer}. However, functions exist that
can be represented compactly with a neural network with $k$ hidden layers but
that would require exponential size (with respect to input size) networks of
depth $k-1$~\citep{haastad1991power, leroux2010deep}.  Therefore, in practice it
is often more convenient to use a larger number of hidden layers with fewer
hidden units each. The trade-off between a larger number of hidden layers and
a larger number of hidden units is discussed in~\citet{burgerjmlr2}.

\subsection{Applying MLPs for image denoising} 
To denoise images, we decompose a given noisy image into overlapping patches.
We then normalize the patches by subtracting $0.5$ and dividing by $0.2$,
denoise each patch separately and perform the inverse normalization (multiply
with $0.2$, add $0.5$) on the denoised patches.
The denoised image is obtained by placing the denoised patches at the locations
of their noisy counterparts, then averaging on the overlapping regions. We
found that we could improve results slightly by weighting the denoised patches
with a Gaussian window. Also, instead of using all possible overlapping patches
(stride size $1$, or patch offset $1$), we found that results were almost
equally good by using every third sliding-window patch (stride size $3$), while
decreasing computation time by a factor of $9$. Using a stride size of $3$, we
were able to denoise images of size $350\times500$ pixels in approximately one
minute (on CPU), which is slower than BM3D~\citep{dabov2007image}, but much
faster than KSVD~\citep{aharon2006rm} and NLSC~\citep{mairal2010non} and also
faster than EPLL~\citep{zoran2011learning}.

\subsection{Efficient implementation on GPU}
The computationally most intensive operations in an MLP are the matrix-vector
multiplications. For these operations \emph{Graphics Processing Units} (GPUs)
are better suited than \emph{Central Processing Units} (CPUs) because of their
ability to efficiently parallelize operations. For this reason we implemented
our MLP on a GPU. We used nVidia's C2050 GPU and achieved a speed-up factor of
more than one order of magnitude compared to an implementation on a quad-core
CPU. This speed-up is a crucial factor, allowing us to run larger-scale
experiments. We describe training for various setups in~\citet{burgerjmlr2}.

\section{Experimental setup}
We performed all our experiments on gray-scale images. These were obtained from
color images with \textsc{matlab}'s \texttt{rbg2gray} function.  Since it is
unlikely that two noise samples are identical, the amount of training data is
effectively infinite, no matter which dataset is used. However, the number of
uncorrupted patches is restricted by the size of the dataset.  Note that the
MLPs could be also trained on color images, possibly exploiting structure
between the different color channels.  However, in this publication we focus on
the gray-scale case.
\medskip

\paragraph{Training data:} 
\label{sec:trainingdata} 
For almost all our experiments, we used images from the imagenet dataset
\citep{imagenet_cvpr09}. Imagenet is a hiearchically organized image database,
in which each node of the hierarchy is depicted by hundreds and thousands of
images. We completely disregard all labels provided with the dataset. We used
$1846296$ images from $2500$ different object categories. We performed no
pre-processing other than the transform to grey-scale on the training images.  
\medskip

\paragraph{Test data:}
\label{sec:testdata}
We define six different test sets to evaluate our approach:
\begin{enumerate}
  \item \textit{standard test images:} This set of $11$ images contains
  standard images, such as ``Lena'' and ``Barbara'', that have been used to
  evaluate other denoising algorithms~\citep{dabov2007image}.
  \item \textit{Berkeley BSDS500:} We used all $500$ images of this dataset as
  a test set.  Subsets of this dataset have been used as a training set for
  other methods such as FoE~\citep{roth2009fields} and
  EPLL~\citep{zoran2011learning}.
  \item \textit{Pascal VOC 2007:} We randomly selected $500$ images from the
  Pascal VOC 2007 test set~\citep{pascal-voc-2007}. 
  \item \textit{Pascal VOC 2011:} We randomly selected $500$ images from the
  Pascal VOC 2011 training set.
  \item \textit{McGill:} We randomly selected $500$ images from the McGill
  dataset~\citep{olmos2004biologically}.
  \item \textit{ImageNet:} We randomly selected 500 images from the ImageNet
  dataset not present in the training set. We also used object categories not
  used in the training set.
\end{enumerate}
We selected dataset 1) because it has become a standard test
dataset, see \citet{dabov2007image} and \citet{mairal2010non}. The images contained in it
are well-known and diverse: Image ``Barbara'' contains a lot of regular
structure, whereas image ``Man'' contains more irregular structure and image
``Lena'' contains smooth areas. We chose to make a more thorough comparison,
which is why we evaluated our approach as well as competing algorithms on five
larger test sets. We chose five different image sets of $500$ images instead of
one set of $2500$ images in order to see if the performance of methods is
significantly affected by the choice of the dataset.
EPLL~\citep{zoran2011learning} is trained on a subset of dataset 2),
NLSC~\citep{mairal2010non} is trained on a subset of 4) and our method is
trained on images extracted from the same larger dataset as 6).

\paragraph{Types of noise:} 
For most of our experiments, we used AWG noise with $\sigma=25$. However, we
also show results for other noise levels.
Finally, we trained MLPs to remove mixed Gaussian-Poisson noise, JPEG artifacts, salt and
pepper noise and noise that resembles stripes.

\section{Results: comparison with existing algorithms}
\label{sec:results}
In this section, we present results achieved with an MLP on AWG noise with
five different noise levels. We also present results achieved on less
well-studied forms of noise. We present in more detail what steps we
took to achieve these results in \citet{burgerjmlr2}.

We compare against the following algorithms:
\begin{enumerate}
  \item KSVD~\citep{aharon2006rm}: This is a dictionary-based method where the
  dictionary is adapted to the noisy image at hand. A noisy patch is denoised
  by approximating it with a sparse linear combination of dictionary elements.
  \item EPLL~\citep{zoran2011learning}: The distribution of image patches is
  described by a mixture of Gaussians. The method presents a novel approach to
  denoising whole images based on patch-based priors. The method was shown to
  be sometimes superior to BM3D~\citep{dabov2007image}, which is often
  considered the state-of-the-art in image denoising.
  \item BM3D~\citep{dabov2007image}: The method does not explicitly use an image
  prior, but rather exploits the fact that images often contain
  self-similarities. Concretely, the method relies on a ``block matching''
  procedure: Patches within the noisy image that are similar to a reference
  patch are denoised together. This approach has been shown to be very
  effective and is often considered the state-of-the-art in image denoising.
  \item NLSC~\citep{mairal2010non}: This is a dictionary-based method which
  (like KSVD) adapts the dictionary to the noisy image at hand. In addition,
  the method exploits image self-similarities, using a block-matching approach
  similar to BM3D. This method also achieves excellent results.
\end{enumerate}
We choose these algorithms for our comparison because they achieve good
results. BM3D and NLSC are usually referred to as the state-of-the-art in image
denoising. Of the four algorithms, KSVD achieves the least impressive results,
but these are still usually better than those achieved with
BLSGSM~\citep{portilla2003image}, which was considered state-of-the-art before
the introduction of KSVD. An additional reason for the choice of these
algorithms is the diversity of the approaches. Learning-based approaches are
represented through EPLL, whereas engineered approaches that don't rely on
learning are represented by BM3D. Non-local methods are represented by BM3D and
NLSC. Finally, dictionary-based approaches are represented by KSVD and NLSC.

\begin{table}
  \centering
  \begin{tabular}{l||ccccc}
    image &  KSVD & EPLL & BM3D & NLSC & MLP \\  
    %image & KSVD\citep{aharon2006rm} & EPLL\citep{zoran2011learning} & BM3D\citep{dabov2007image} & NLSC\citep{mairal2010non} & MLP \\
    \hline\hline 
    Barbara &  29.49dB  &  28.52dB  &  \textbf{30.67}dB  &  \textit{30.50}dB  &  29.52dB  \\  
    Boat &  29.24dB  &  29.64dB  &  29.86dB  &  \textit{29.86}dB  &  \textbf{29.95}dB  \\  
    C.man &  28.64dB  &  29.18dB  &  29.40dB  &  \textit{29.46}dB  &  \textbf{29.60}dB  \\  
    Couple &  28.87dB  &  29.45dB  &  \textit{29.68}dB  &  29.63dB  &  \textbf{29.75}dB  \\  
    F.print &  27.24dB  &  27.11dB  &  \textbf{27.72}dB  &  27.63dB  &  \textit{27.67}dB  \\  
    Hill &  29.20dB  &  29.57dB  &  \textit{29.81}dB  &  29.80dB  &  \textbf{29.84}dB  \\  
    House &  32.08dB  &  32.07dB  &  \textit{32.92}dB  &  \textbf{33.08}dB  &  32.52dB  \\  
    Lena &  31.30dB  &  31.59dB  &  \textit{32.04}dB  &  31.87dB  &  \textbf{32.28}dB  \\  
    Man &  29.08dB  &  29.58dB  &  29.58dB  &  \textit{29.62}dB  &  \textbf{29.85}dB  \\  
    Montage &  30.91dB  &  31.18dB  &  \textbf{32.24}dB  &  \textit{32.15}dB  &  31.97dB  \\  
    Peppers &  29.69dB  &  30.08dB  &  30.18dB  &  \textbf{30.27}dB  &  \textbf{30.27}dB  \\  
  \end{tabular}
  \caption{Results on $11$ standard test images for $\sigma=25$.}
  \label{tab:results001}
\end{table}

\begin{figure*}
  \centering
  %\begin{tabular}{cccc}
  \begin{tabular}{ccc}
    \includegraphics[width=0.3\textwidth]{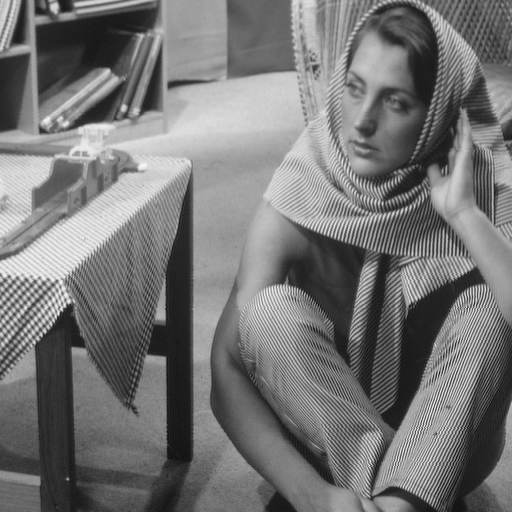} &
    \includegraphics[width=0.3\textwidth]{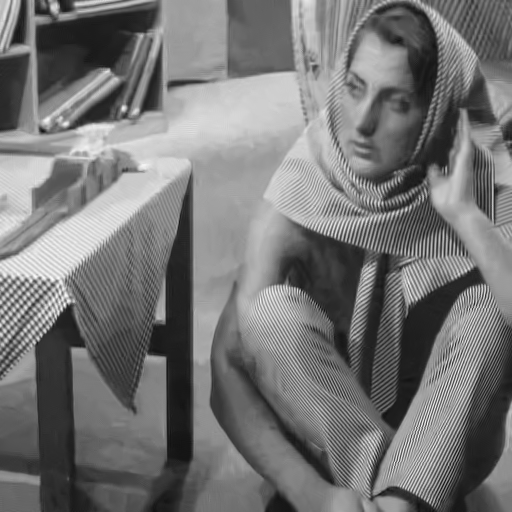} & 
    \includegraphics[width=0.3\textwidth]{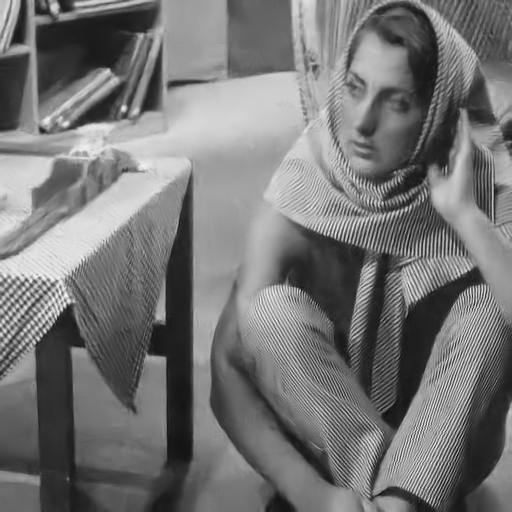} \\
    clean image (Barbara) & BM3D: $\mathbf{30.67}$dB & MLP: $29.52$dB \\
    % From the VOC 2007 test set:
    \includegraphics[width=0.3\textwidth]{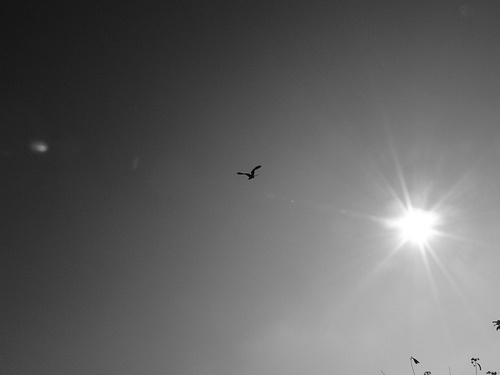} &
    \includegraphics[width=0.3\textwidth]{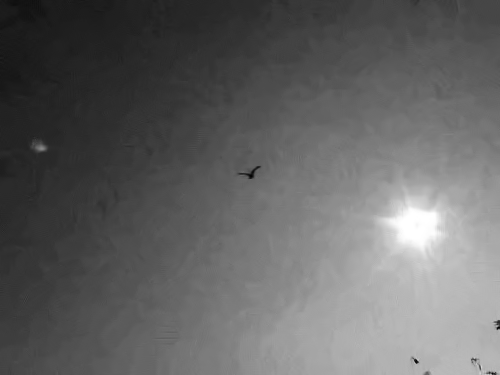} & 
    \includegraphics[width=0.3\textwidth]{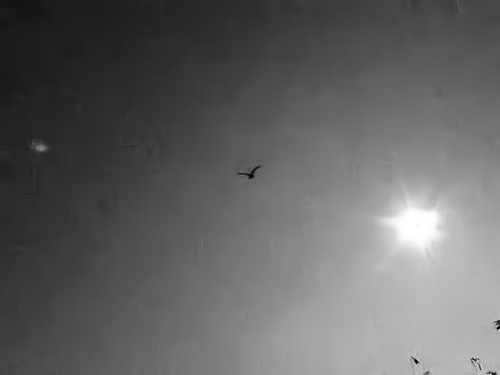} \\
    clean image (004513) & BM3D: $38.92$dB & MLP: $\mathbf{40.57}$dB\\
    % From the BSDS500 set:
    \includegraphics[width=0.3\textwidth]{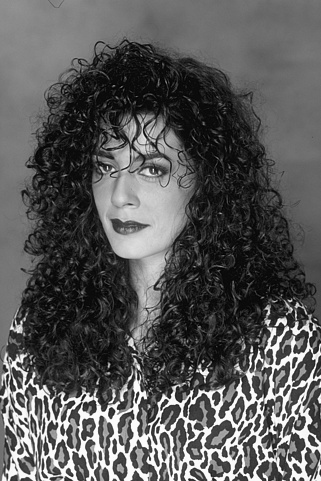} &
    \includegraphics[width=0.3\textwidth]{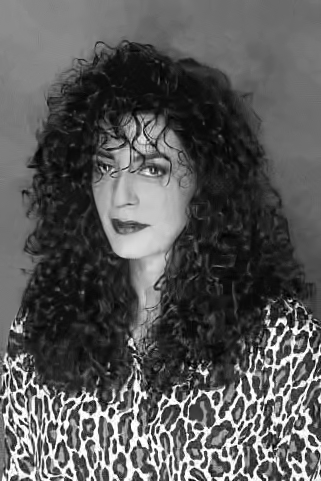} & 
    \includegraphics[width=0.3\textwidth]{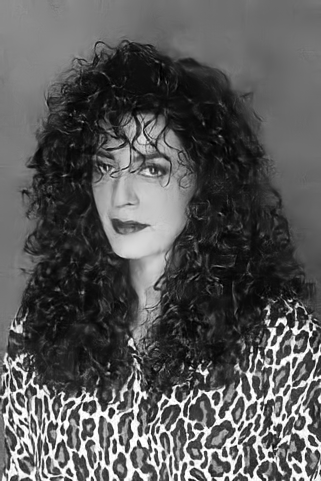} \\
    clean image (198054) & BM3D: $26.28$dB & MLP: $\mathbf{27.09}$dB
  \end{tabular}
  \caption{We outperform BM3D on images with smooth surfaces and non-regular structures. BM3D outperforms us on images with regular structure. The image ``Barbara'' contains a lot of regular structure on the pants as well the table-cloth.}
  \label{fig:imagecomparison}
\end{figure*}

\subsection{Detailed comparison on one noise level}
We will now compare the results achieved with an MLP to results achieved with
other denoising algorithms on AWG noise with $\sigma=25$. We choose the MLP
with architecture ($39\times2, 3072, 3072, 2559, 2047, 17\times2$) because it
delivered the best results.  The MLP was trained for approximately $3.5 \cdot
10^8$ backprops, see \citet{burgerjmlr2} for details.

\paragraph{Comparison on 11 standard test images:}
Table~\ref{tab:results001} summarizes the comparison of our approach (MLP) to
the four other denoising algorithms. Our approach achieves the best result on
$7$ of the $11$ test images and is the runner-up on one image.  However, our
method is clearly inferior to BM3D and NLSC on images ``Barbara'' and
``House''. These two images contain a lot of regular structure (see
Figure~\ref{fig:imagecomparison}) and are therefore ideally suited for
algorithms like BM3D and NLSC, which adapt to the noisy image. However, we
outperform KSVD on both of these images even though KSVD is also an algorithm
that is well-suited for these types of images.  We also note that we outperform
both KSVD and EPLL on every image.

\begin{figure}[htbp]
  \centering
  \begin{tabular}{cc}
    \includegraphics[width=0.45\textwidth]{images_im_201_BM3D_berkeley500-sigma50.png} &
    \includegraphics[width=0.45\textwidth]{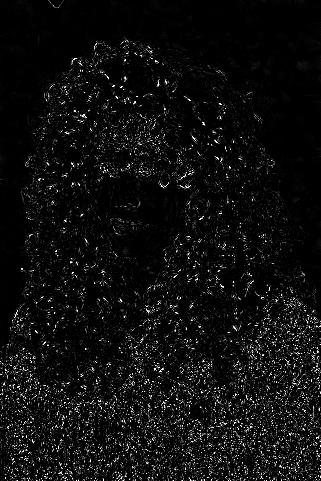} \\
    (a) & (b) 
  \end{tabular}
  \caption{The MLP outperforms BM3D on image (a). Locations where BM3D
  is worse than the MLP on image 198054 are highlighted (b).}
  \label{fig:errormap}
\end{figure}

\begin{figure}[htbp]
  \centering
    \includegraphics[width=\columnwidth]{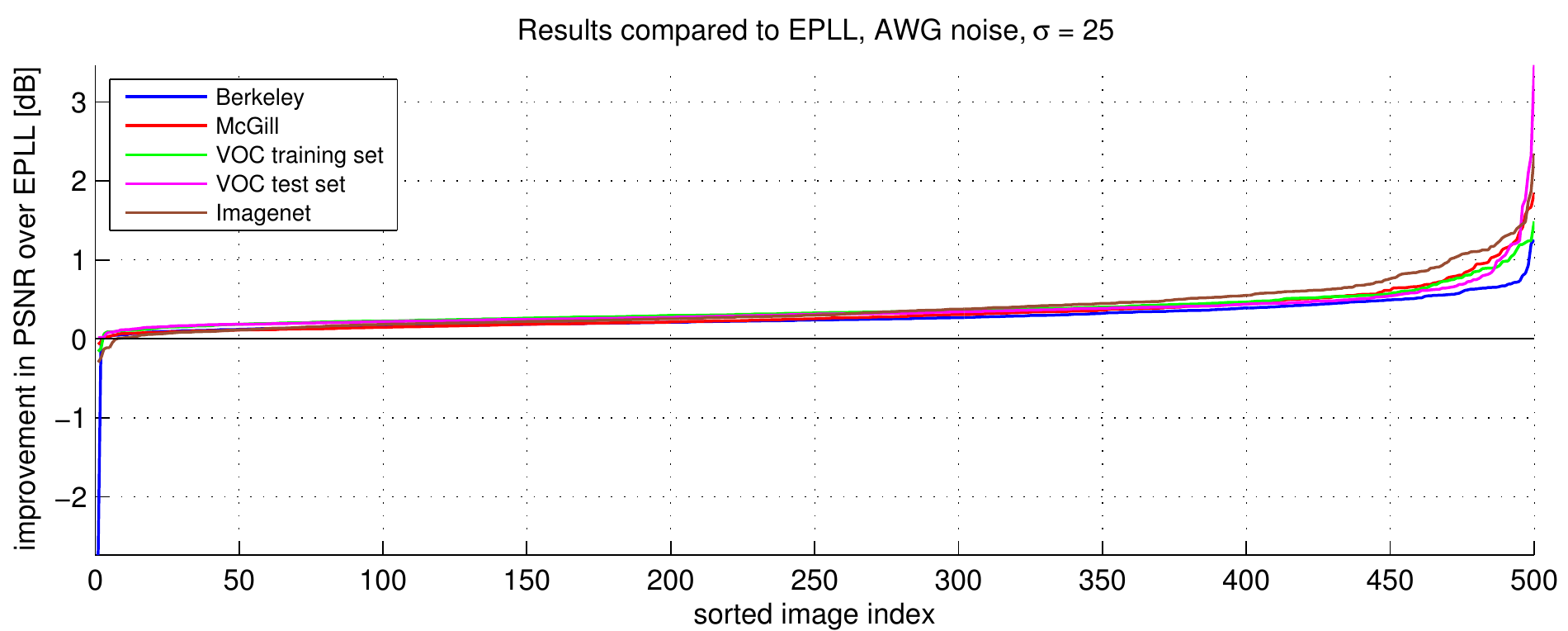}
    \includegraphics[width=\columnwidth]{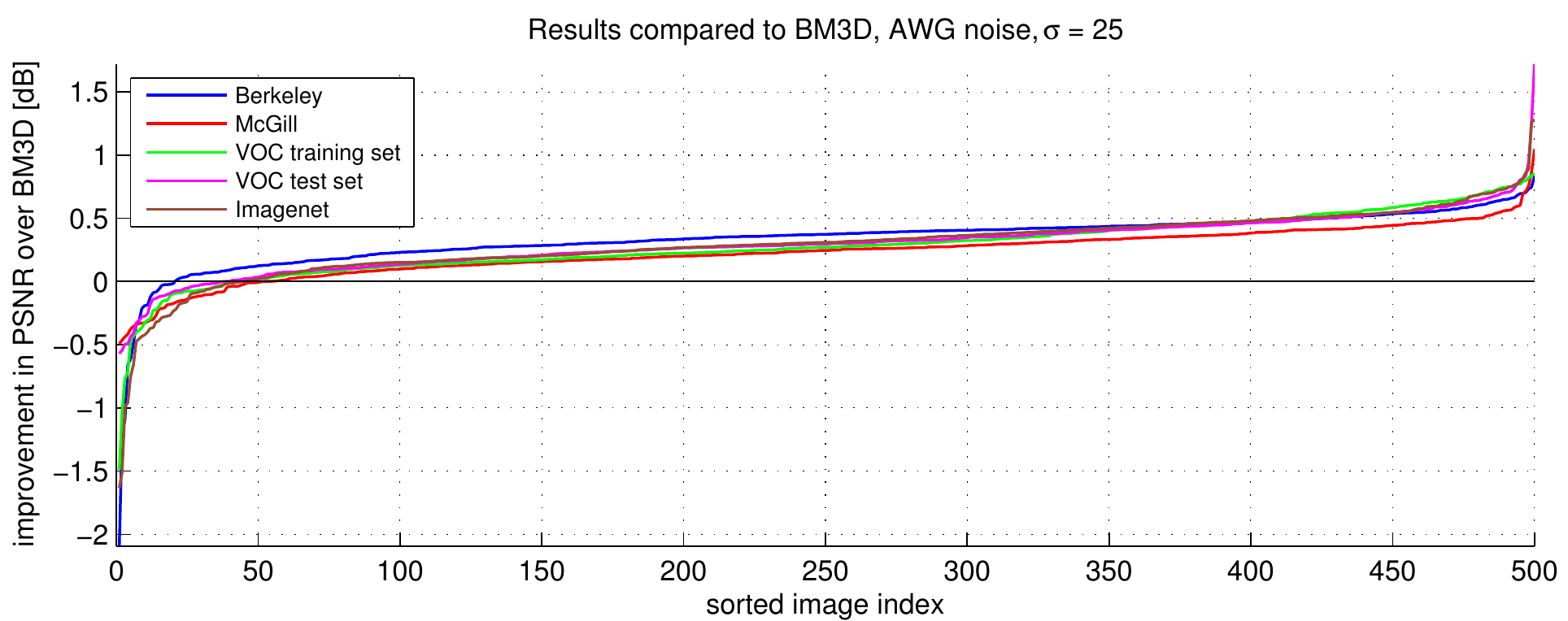}
    \includegraphics[width=\columnwidth]{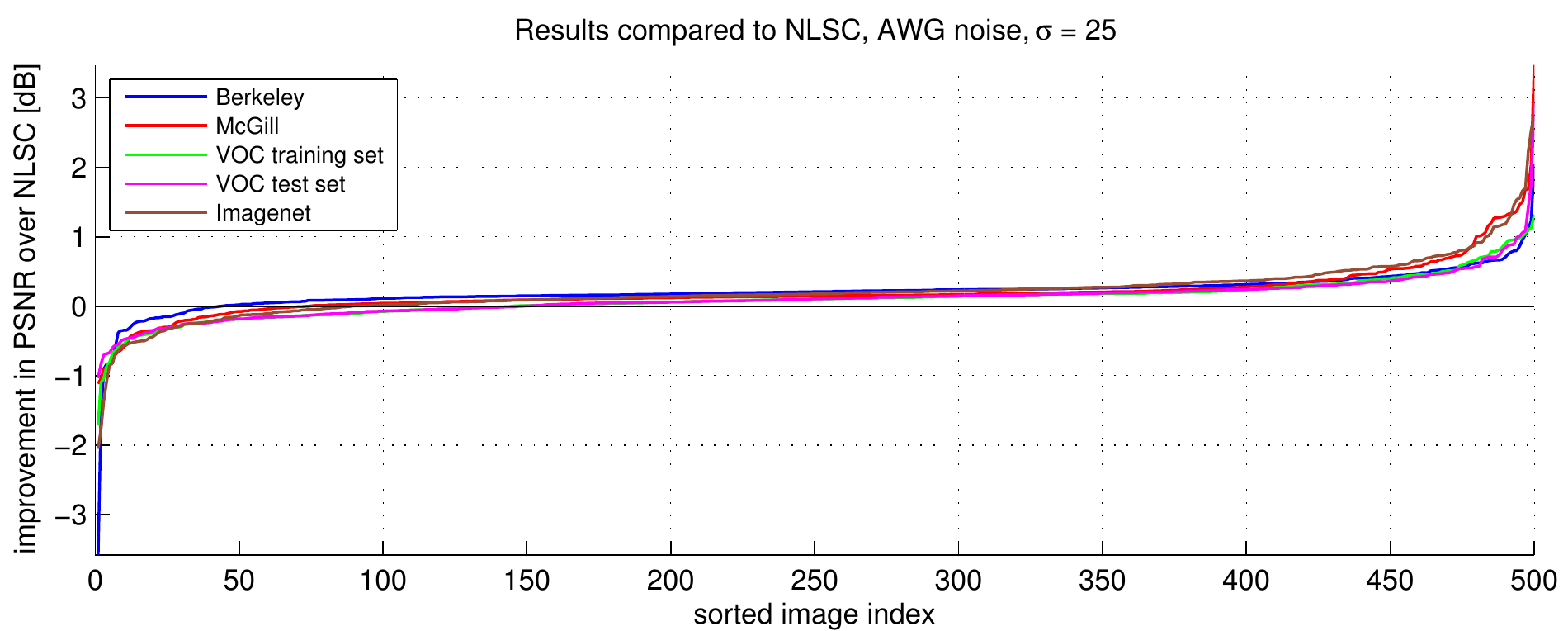}
    \caption{Results compared to EPLL (top), BM3D (middle) and NLSC (bottom) on five datasets of 500 images, $\sigma=25$.}
  \label{fig:comparison}
\end{figure}

\paragraph{Comparison on larger test sets}
We now compare our approach to EPLL, BM3D and NLSC on the five larger test sets
defined in section~\ref{sec:testdata}. Each dataset contains $500$ images,
giving us a total of $2500$ test images.

\begin{itemize}
\item Comparison to EPLL: We outperform EPLL on $2487$ ($99.48\%$) of the
$2500$ images, see Figure~\ref{fig:comparison}. The average improvement over
all datasets is $0.35$dB. On the VOC 2007 test set, we outperform EPLL on
every image. The best average improvement over EPLL was on the subset of the
ImageNet dataset ($0.39$dB), whereas the smallest improvement was on the
Berkeley dataset ($0.27$dB). This is perhaps a reflection of the fact that EPLL
was trained on a subset of the Berkeley dataset, whereas our approach was
trained on the ImageNet dataset. For EPLL, the test set contains the training
set. For our method, this is not the case, but it is plausible that the
ImageNet dataset contains some form of regularity across the whole dataset.

\item Comparison to BM3D: We outperform BM3D on $2304$ ($92.16\%$) of the
$2500$ images, see Figure~\ref{fig:comparison}. The average improvement over
all datasets is $0.29$dB. The largest average improvement was on the Berkeley
dataset ($0.34$dB), whereas the smallest average improvement was on the McGill
dataset ($0.23$dB).

Figure~\ref{fig:errormap} highlights the areas of the image in the lower row of
Figure~\ref{fig:imagecomparison} where BM3D creates larger errors than the MLP.
We see that it is indeed in the areas with complicated structures (the
hair and the shirt) that the MLP has an advantage over BM3D.

\item Comparison to NLSC: We outperform NLSC on $2003$ ($80.12\%$) of the
$2500$ images, see Figure~\ref{fig:comparison}. The average improvement over
all datasets was $0.16$dB. The largest average improvements were on the
ImageNet subset and Berkeley dataset ($0.21$dB), whereas the smallest average
improvements were on the VOC 2011 training set and VOC 2007 test set ($0.10$dB
and $0.11$dB respectively). This is perhaps a reflection of the fact that the
initial dictionary of  NLSC was trained on a subset of the VOC 2007
dataset~\citep{mairal2010non}.
\end{itemize}

In summary, our method outperforms state-of-the-art denoising algorithms
for AWG noise with $\sigma=25$. The improvement is consistent across datasets.
We notice that our method tends to outperform BM3D on images with smooth areas
such as the sky and on images which contain irregular structure, such as the
hair of the woman in Figure~\ref{fig:imagecomparison}. The fact that our method
performs well on smooth surfaces can probably be explained by the fact that our
method uses large input patches: This allows our method to handle low
frequency noise. Methods using smaller patches (such as BM3D) are blind to
lower frequencies. The fact that our method performs better than BM3D on images
with irregular structures is explained by the block-matching approach employed
by BM3D: The method cannot find similar patches in images with irregular
textures.

\subsection{Comparison on different noise variances}
We have seen that our method achieves state-of-the-art results on AWG noise
with $\sigma=25$. We now evaluate our approach on other noise levels. We use
$\sigma=10$ (low noise), $\sigma=50$ (high noise), $\sigma=75$ (very high
noise) and $\sigma=170$ (extremely high noise) for this purpose. We describe in
\citet{burgerjmlr2} which architectures and patch sizes are used for the various
noise levels.

\begin{table}[htbp]
  \centering
  \begin{tabular}{l||ccccc}
    %image & KSVD\citep{aharon2006rm} & EPLL\citep{zoran2011learning} & BM3D\citep{dabov2007image} & NLSC\citep{mairal2010non} & MLP \\
    image & KSVD & EPLL & BM3D & NLSC & MLP \\
    \hline\hline 
    Barbara &  34.40dB  &  33.59dB  &  \textbf{34.96}dB  &  \textbf{34.96}dB  &  34.07dB  \\  
    Boat &  33.65dB  &  33.64dB  &  \textit{33.89}dB  &  \textbf{34.02}dB  &  33.85dB  \\  
    C.man &  33.66dB  &  33.99dB  &  34.08dB  &  \textbf{34.15}dB  &  \textit{34.13}dB  \\  
    Couple &  33.51dB  &  33.82dB  &  \textbf{34.02}dB  &  \textit{33.98}dB  &  33.89dB  \\  
    F.print &  32.39dB  &  32.12dB  &  32.46dB  &  \textit{32.57}dB  &  \textbf{32.59}dB  \\  
    Hill &  33.37dB  &  33.49dB  &  \textit{33.60}dB  &  \textbf{33.66}dB  &  33.59dB  \\  
    House &  35.94dB  &  35.74dB  &  \textit{36.71}dB  &  \textbf{36.90}dB  &  35.94dB  \\  
    Lena &  35.46dB  &  35.56dB  &  \textbf{35.92}dB  &  35.85dB  &  \textit{35.88}dB  \\  
    Man &  33.53dB  &  33.94dB  &  33.97dB  &  \textit{34.06}dB  &  \textbf{34.10}dB  \\  
    Montage &  35.91dB  &  36.45dB  &  \textbf{37.37}dB  &  \textit{37.24}dB  &  36.51dB  \\  
    Peppers &  34.20dB  &  34.54dB  &  34.69dB  &  \textbf{34.78}dB  &  \textit{34.72}dB
  \end{tabular}
  \caption{Results on $11$ standard test images for $\sigma=10$.}
  \label{tab:results002}
\end{table}
\begin{table}[htbp]
  \centering
  \begin{tabular}{l||ccccc}
    %image & KSVD\citep{aharon2006rm} & EPLL\citep{zoran2011learning} & BM3D\citep{dabov2007image} & NLSC\citep{mairal2010non} & MLP \\
    image & KSVD & EPLL & BM3D & NLSC & MLP \\
    \hline\hline 
    Barbara &  25.22dB  &  24.83dB  &  \textbf{27.21}dB  &  \textit{27.13}dB  &  25.37dB  \\  
    Boat &  25.90dB  &  26.59dB  &  26.72dB  &  \textit{26.73}dB  &  \textbf{27.02}dB  \\  
    C.man &  25.42dB  &  26.05dB  &  26.11dB  &  \textit{26.36}dB  &  \textbf{26.42}dB  \\  
    Couple &  25.40dB  &  26.24dB  &  \textit{26.43}dB  &  26.33dB  &  \textbf{26.71}dB  \\  
    F.print &  23.24dB  &  23.59dB  &  \textbf{24.53}dB  &  \textit{24.25}dB  &  24.23dB  \\  
    Hill &  26.14dB  &  26.90dB  &  \textit{27.14}dB  &  27.05dB  &  \textbf{27.32}dB  \\  
    House &  27.44dB  &  28.77dB  &  \textit{29.71}dB  &  \textbf{29.88}dB  &  29.52dB  \\  
    Lena &  27.43dB  &  28.39dB  &  \textit{28.99}dB  &  28.88dB  &  \textbf{29.34}dB  \\  
    Man &  25.83dB  &  26.68dB  &  \textit{26.76}dB  &  26.71dB  &  \textbf{27.08}dB  \\  
    Montage &  26.42dB  &  27.13dB  &  27.69dB  &  \textit{28.02}dB  &  \textbf{28.07}dB  \\  
    Peppers &  25.91dB  &  26.64dB  &  26.69dB  &  \textit{26.73}dB  &  \textbf{26.74}dB
  \end{tabular}
  \caption{Results on $11$ standard test images for $\sigma=50$.}
  \label{tab:results003}
\end{table}
\begin{table}[htbp]
  \centering
  \begin{tabular}{l||ccccc}
    %image & KSVD\citep{aharon2006rm} & EPLL\citep{zoran2011learning} & BM3D\citep{dabov2007image} & NLSC\citep{mairal2010non} & MLP \\
    image & KSVD & EPLL & BM3D & NLSC & MLP \\
    \hline\hline 
    Barbara &  22.65dB  &  22.95dB  &  \textbf{25.10}dB  &  \textit{25.03}dB  &  23.48dB  \\  
    Boat &  23.59dB  &  24.86dB  &  \textit{25.04}dB  &  24.95dB  &  \textbf{25.43}dB  \\  
    C.man &  23.04dB  &  24.19dB  &  \textit{24.37}dB  &  24.24dB  &  \textbf{24.72}dB  \\  
    Couple &  23.43dB  &  24.46dB  &  \textit{24.71}dB  &  24.48dB  &  \textbf{25.09}dB  \\  
    F.print &  20.72dB  &  21.44dB  &  \textbf{22.83}dB  &  \textit{22.48}dB  &  22.41dB  \\  
    Hill &  24.21dB  &  25.42dB  &  \textit{25.60}dB  &  25.57dB  &  \textbf{25.97}dB  \\  
    House &  24.53dB  &  26.69dB  &  27.46dB  &  \textit{27.64}dB  &  \textbf{27.75}dB  \\  
    Lena &  24.87dB  &  26.50dB  &  27.16dB  &  \textit{27.17}dB  &  \textbf{27.66}dB  \\  
    Man &  23.76dB  &  25.07dB  &  \textit{25.29}dB  &  25.15dB  &  \textbf{25.63}dB  \\  
    Montage &  23.58dB  &  24.86dB  &  \textit{25.36}dB  &  25.20dB  &  \textbf{25.93}dB  \\  
    Peppers &  23.09dB  &  24.52dB  &  \textit{24.71}dB  &  24.46dB  &  \textbf{24.87}dB
  \end{tabular}
  \caption{Results on $11$ standard test images for $\sigma=75$.}
  \label{tab:results004}
\end{table}
\begin{table}[htbp]
  \centering
  \begin{tabular}{l||ccccc}
    %image & KSVD\citep{aharon2006rm} & EPLL\citep{zoran2011learning} & BM3D\citep{dabov2007image} & NLSC\citep{mairal2010non} & MLP \\
    image & KSVD & EPLL & BM3D & NLSC & MLP \\
    \hline\hline 
    Barbara &  18.08dB  &  20.79dB  &  19.74dB  &  \textit{20.99}dB  &  \textbf{21.37}dB  \\  
    Boat &  18.42dB  &  \textit{21.60}dB  &  20.49dB  &  21.48dB  &  \textbf{22.47}dB  \\  
    C.man &  18.00dB  &  20.48dB  &  19.65dB  &  \textit{20.50}dB  &  \textbf{21.28}dB  \\  
    Couple &  18.26dB  &  \textit{21.48}dB  &  20.39dB  &  21.29dB  &  \textbf{22.16}dB  \\  
    F.print &  16.75dB  &  17.06dB  &  17.46dB  &  \textit{18.51}dB  &  \textbf{18.57}dB  \\  
    Hill &  18.69dB  &  \textit{22.63}dB  &  20.98dB  &  22.62dB  &  \textbf{23.33}dB  \\  
    House &  18.20dB  &  \textit{22.52}dB  &  21.19dB  &  21.95dB  &  \textbf{23.80}dB  \\  
    Lena &  18.68dB  &  22.96dB  &  21.38dB  &  \textit{23.20}dB  &  \textbf{24.24}dB  \\  
    Man &  18.49dB  &  \textit{22.10}dB  &  20.59dB  &  21.72dB  &  \textbf{22.85}dB  \\  
    Montage &  17.91dB  &  \textit{20.48}dB  &  19.69dB  &  20.40dB  &  \textbf{20.93}dB  \\  
    Peppers &  17.47dB  &  \textit{20.26}dB  &  19.58dB  &  19.53dB  &  \textbf{20.81}dB
  \end{tabular}
  \caption{Results on $11$ standard test images for $\sigma=170$.}
  \label{tab:results005}
\end{table}

\paragraph{Comparison on 11 standard test images:}
Table~\ref{tab:results002} compares our method against KSVD, EPLL, BM3D and
NLSC on the test set of $11$ standard test images for $\sigma=10$. Our method
outperforms KSVD on ten images, EPLL on all images, BM3D on four images and
NLSC on three images. Our method achieves the best result of all algorithms on
two images.  Like for $\sigma=25$, BM3D and NLSC perform particularly well for
images ``Barbara'' and ``House''.

Table~\ref{tab:results003} performs the same comparison for $\sigma=50$. Our
method outperforms all others on $8$ of the $11$ images. BM3D and NLSC still
perform significantly better on the image ``Barbara''. We outperform KSVD and
EPLL on every image.

For $\sigma=75$, our method outperforms all others on $9$ of the $11$ images,
see Table~\ref{tab:results004}. BM3D and NLSC still perform significantly
better on the image ``Barbara''.

For $\sigma=170$, our method outperforms all other methods on all images, see
Table~\ref{tab:results005}. It was suggested by \citet{levin2010natural} that
image priors are not useful at extremely high noise levels. However, our
results suggest otherwise: Our method is the best-performing method on this
noise level. The second best performing method, EPLL, is also a prior-based
method.  The improvement of our method over BM3D (which is not prior-based) is
often very high (almost $3$dB on image ``Lena''). 

\begin{figure}[htbp]
  \centering
    \includegraphics[width=\columnwidth]{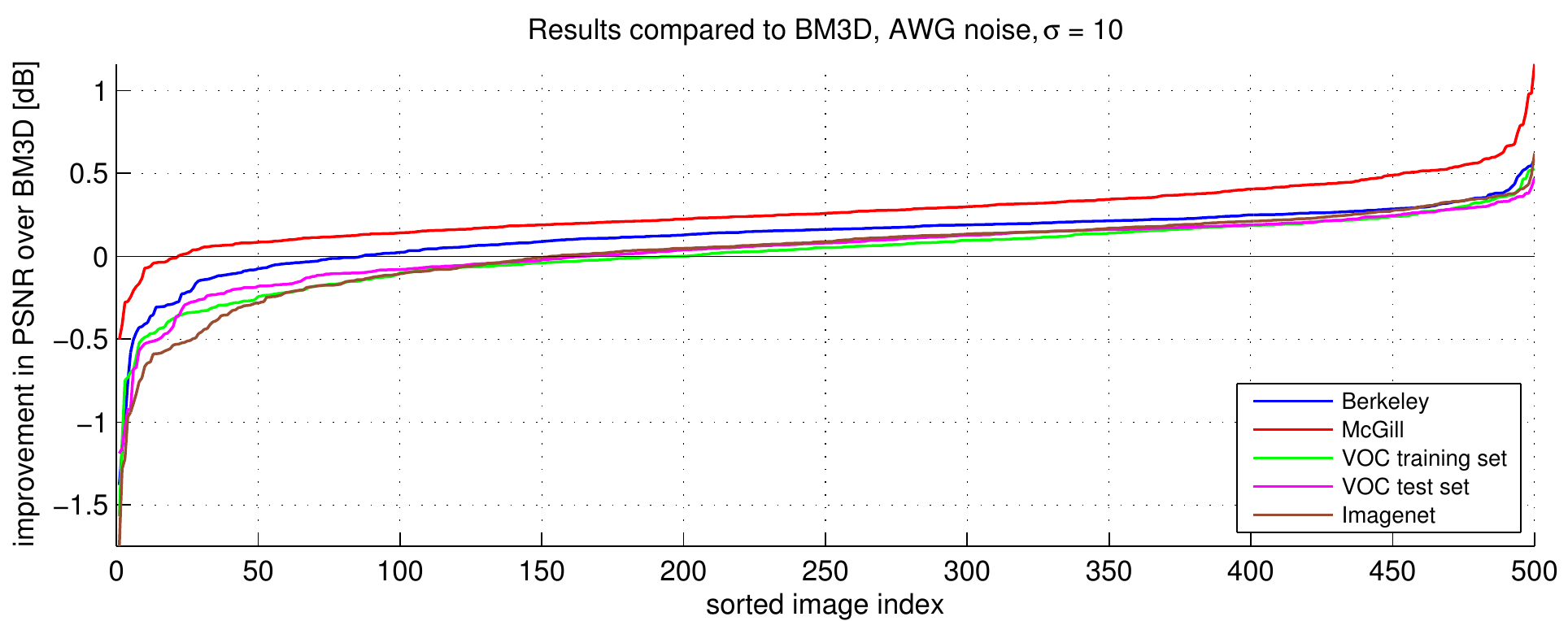}
    \includegraphics[width=\columnwidth]{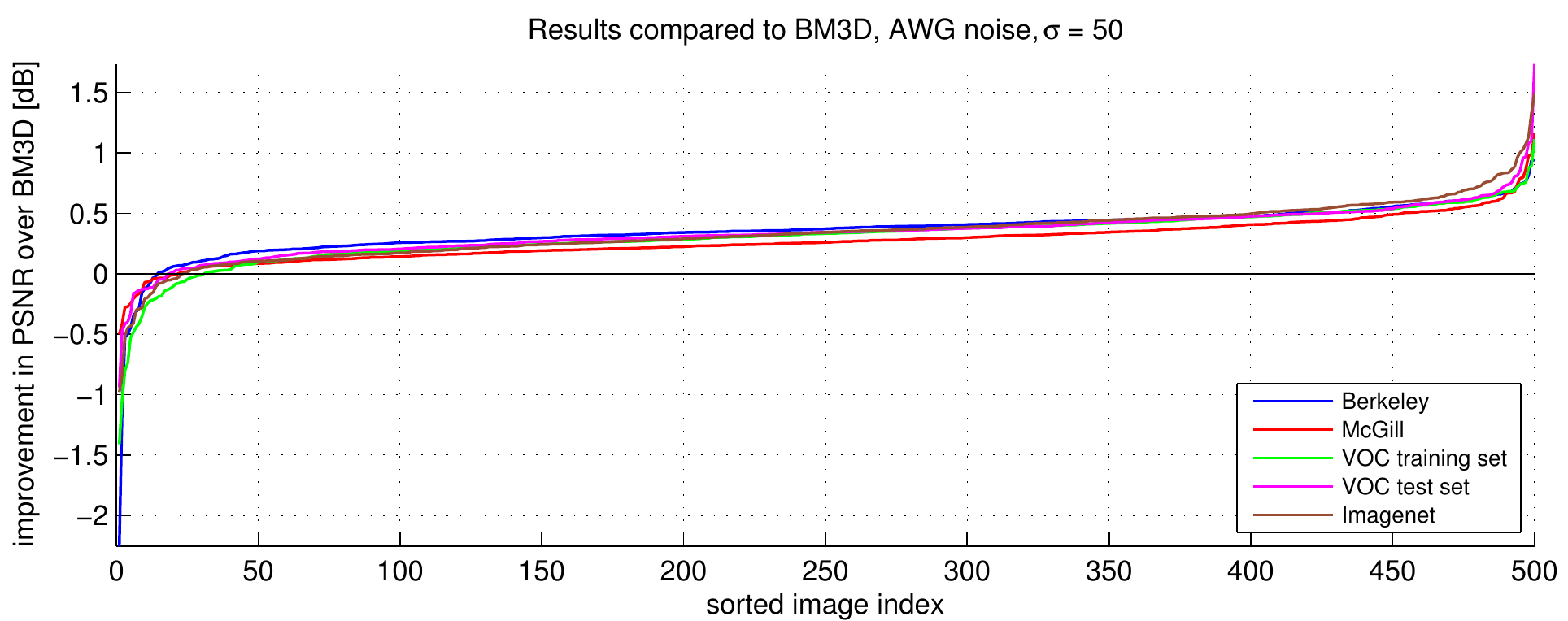}
    \includegraphics[width=\columnwidth]{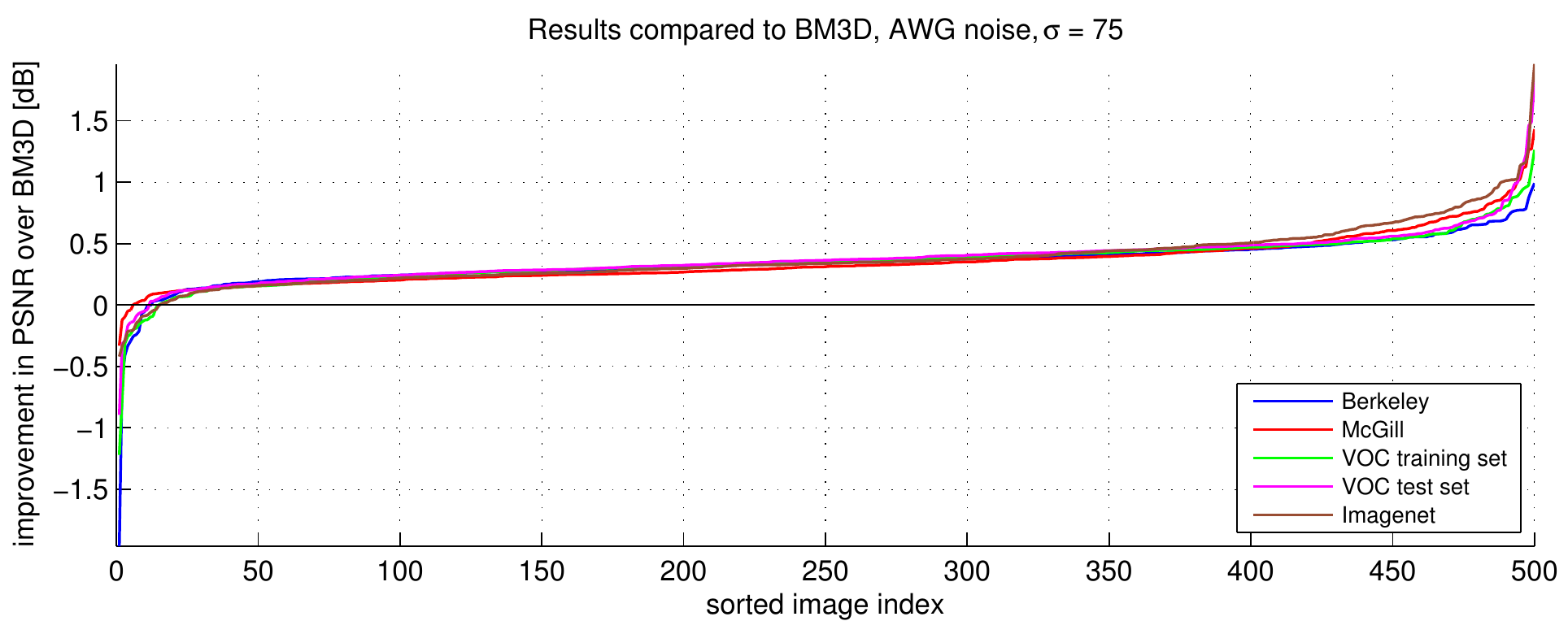}
    \caption{Results compared to BM3D on five datasets of 500 images and different noise levels. Top: $\sigma=10$, middle: $\sigma=50$, bottom: $\sigma=75$.}
  \label{fig:comparison2}
\end{figure}

\begin{figure}[t]
  \centering
    \includegraphics[width=\columnwidth]{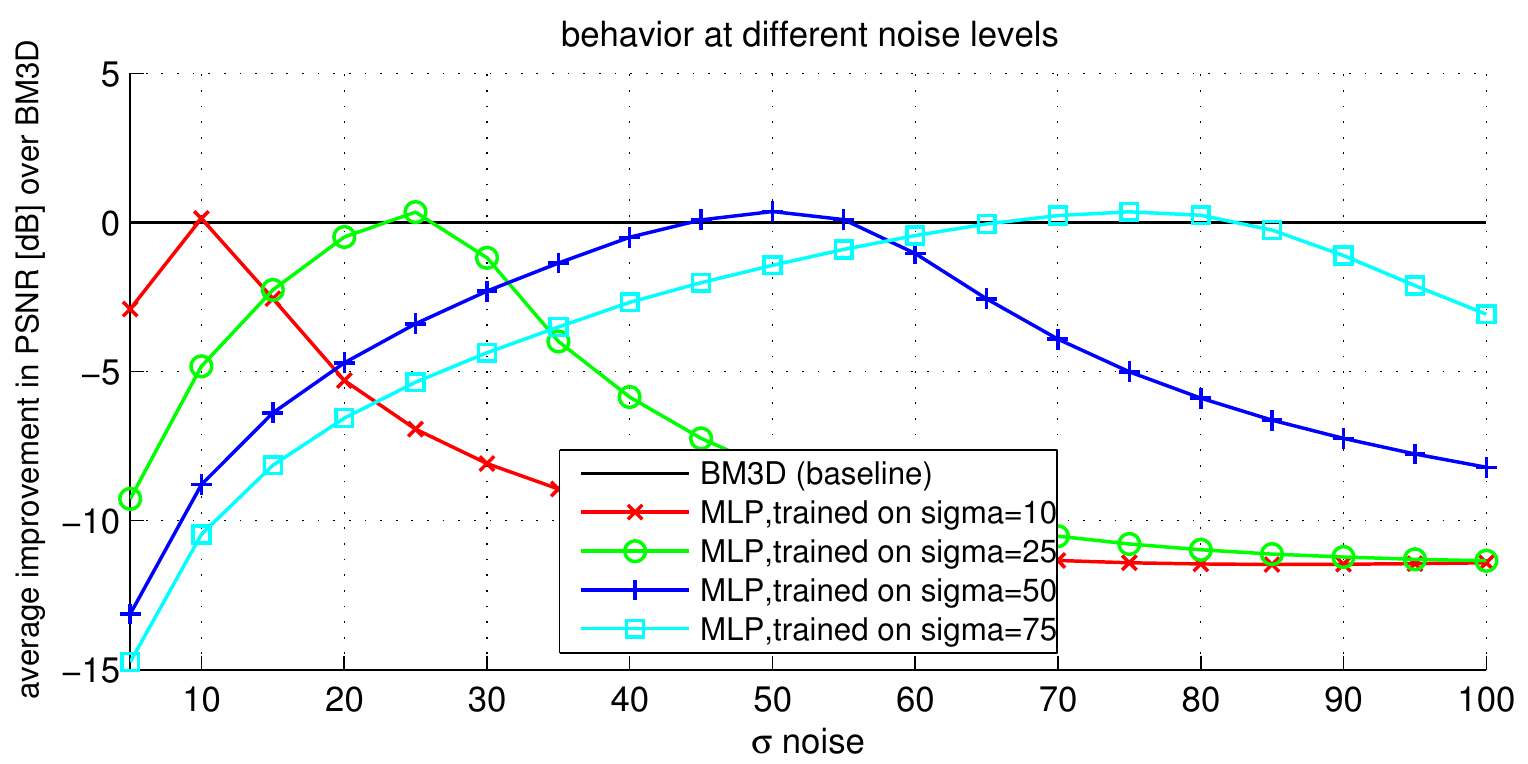}
    \caption{Results achieved on different noise levels. Results are averaged over the $500$ images
      in the Berkeley dataset.}
  \label{fig:manysigmas001}
\end{figure}

\paragraph{Comparison on 2500 test images:}
Figure~\ref{fig:comparison2} (top) compares the results achieved with an MLP on
$\sigma=10$ to BM3D. We outperform BM3D on $1876$ ($75.04\%$) of the $2500$
images. The average improvement over all images is $0.1$dB. The largest average
improvement is on the McGill dataset ($0.27$dB), whereas the smallest average
improvement is on the VOC training set ($0.02$dB). The improvement in PSNR is
very small on the VOC training set, but we observe an improvement on $301$
($60.2\%$) of the $500$ images.

Figure~\ref{fig:comparison2} (middle) compares the results achieved with an MLP on
$\sigma=50$ to BM3D. We outperform BM3D on $2394$ ($95.76\%$) of the $2500$
images. The average improvement over all datasets is $0.32$dB. The largest
average improvement is on the Berkeley dataset ($0.36$dB), whereas the
smallest average improvement is on the McGill dataset ($0.27$dB). This is an
even greater improvement over BM3D than on $\sigma=25$, see
Figure~\ref{fig:comparison}. 

Figure~\ref{fig:comparison2} (bottom) compares the results achieved with an MLP on
$\sigma=75$ to BM3D. We outperform BM3D on $2440$ ($97.60\%$) of the $2500$
images. The average improvement over all datasets is $0.36$dB.  The average
improvement is almost the same for all datasets, ranging from $0.34$ to
$0.37$dB.

\paragraph{Adaptation to other noise levels:} How do the MLPs perform on
noise levels they have not been trained on?  Figure~\ref{fig:manysigmas001}
summarizes the results achieved by MLPs on noise levels they have not been
trained on and compares these results to BM3D. The results are averaged over
the $500$ images in the Berkeley dataset. We varied $\sigma$ between $5$ and
$100$ in steps of $5$.  We see that the MLPs achieve better results than BM3D
on the noise levels they have been trained on.  However, the performance
degrades quickly for noise levels they have not been trained on. Exceptions are
the MLPs trained on $\sigma=50$ and $\sigma=75$, which also outperform BM3D on
$\sigma=45$ and $\sigma=55$ (for the MLP trained on $\sigma=50$) and
$\sigma=70$ and $\sigma=80$ (for the MLP trained on $\sigma=75$).

We conclude than our method is particularly well suited for medium to high
noise levels. We outperform the previous state-of-the-art on all noise levels,
but for $\sigma=10$, the improvement is rather small ($0.1$dB).  However, our
method has to be trained on each noise level in order to achieve good results.

\clearpage
\section{Results: Comparison with theoretical bounds}
\label{sec:comparisonbounds}
It has been observed that recent denoising algorithms tend to perform
approximately equally well~\citep{chatterjee2010denoising}, which naturally
raises the question of whether recent state-of-the-art algorithms are close to
an inherent limit on denoising quality. Two approaches to estimating bounds on
denoising performance have been followed~\citep{chatterjee2010denoising,
levin2010natural}. We will relate the results obtained by our algorithm to
these bounds.

\begin{figure}[htbp]
  \centering
    \begin{tabular}{cc}
      \includegraphics[width=0.48\columnwidth]{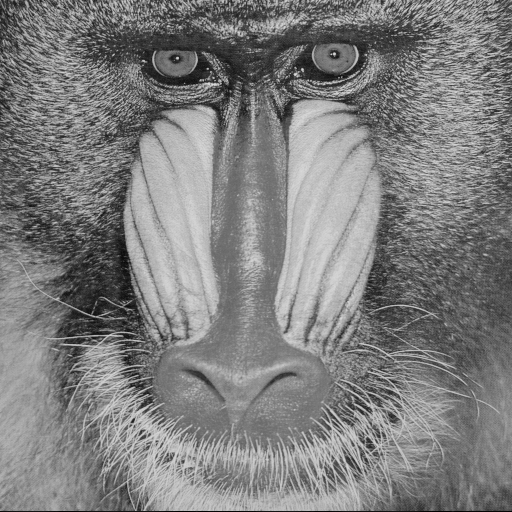} &
      \includegraphics[width=0.48\columnwidth]{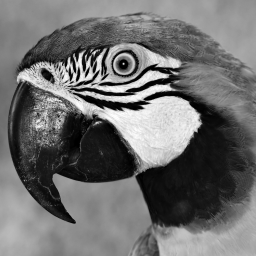}
    \end{tabular}
    \caption{Images ``Mandrill'' and ``Parrot''. For $\sigma=25$, the
    theoretical bounds estimated by~\citep{chatterjee2010denoising} are very
    close to the result achieved by BM3D: $25.61$dB and $28.94$dB,
    respectively. Our results outperform these bounds
    and are $26.01$dB and $29.25$dB
    respectively.}
  \label{fig:mandrill}
\end{figure}
\subsection{Clustering-based bounds} The authors
of~\citep{chatterjee2010denoising} derive bounds on image denoising capability.
The authors make a ``cluster'' assumption about images: Each patch in a noisy
image is assigned to one of a finite number of clusters. Clusters with more
patches are denoised better than clusters with fewer patches.  According to
their bounds, improvements over existing denoising algorithms are mainly to be
achieved on images with simple geometric structure (the authors use a synthetic
``box'' image as an example), whereas current denoising algorithms (and BM3D in
particular) are already very close to the theoretical bounds for images with
richer geometric structure. 

Figure~\ref{fig:mandrill} shows two images with richer structure and on which
BM3D is very close to the estimated theoretical bounds for $\sigma=25$ 
\citep[][Fig.~11]{chatterjee2010denoising}. Very little, if any, improvement is
expected on these images. Yet, we outperform BM3D by $0.4$dB and $0.31$dB on
these images, which is a significant improvement.

The MLP does not operate according to the cluster assumption (it operates on a
single patch at a time) and it performs particularly well on images with rich
geometric structure. We therefore speculate that the cluster assumption might
not be a reasonable assumption to derive ultimate bounds on image denoising
quality.

\begin{table}
  \centering
  \begin{tabular}{l|ccc}
    & worst & best & mean \\  
    \hline 
    BLSGSM~\citep{portilla2003image} & 22.65dB  & 23.57dB  & 23.15dB  \\  
    KSVD~\citep{aharon2006rm} & 21.69dB  & 22.59dB  & 22.16dB  \\  
    NLSC~\citep{mairal2010non} & 21.39dB  & 22.49dB  & 21.95dB  \\  
    BM3D~\citep{dabov2007image} &  \textit{22.94}dB  & 23.96dB  & 23.51dB  \\  
    BM3D, step1~\citep{dabov2007image} & 21.85dB  & 22.79dB  & 22.35dB  \\  
    EPLL~\citep{zoran2011learning} & 22.94dB  &  \textit{24.07}dB  &  \textit{23.56}dB  \\  
    MLP &  \textbf{23.32}dB  &  \textbf{24.34}dB  &  \textbf{23.85}dB
  \end{tabular}
  \caption{Comparison of results achieved by different methods on the
  down-sampled and cropped ``Peppers'' image for $\sigma=75$ and $100$
  different noisy instances.}
  \label{tab:peppersbounds}
\end{table}

\begin{figure}[htbp]
  \centering
    \begin{tabular}{cc}
      \includegraphics[width=0.48\columnwidth]{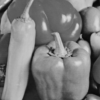} &
      \includegraphics[width=0.48\columnwidth]{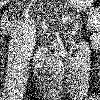} \\
      (a) Original image & (b) Noisy input, $10.57$dB \\
      \includegraphics[width=0.48\columnwidth]{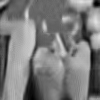} &
      \includegraphics[width=0.48\columnwidth]{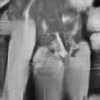} \\
      (c) BM3D, $23.92$dB & (d) MLP, $\mathbf{24.27}$dB 
    \end{tabular}
    \caption{For image (a) and $\sigma=75$, the best achievable result
    estimated in~\citep{levin2010natural} is only $0.07$dB better than the
    result achieved by BM3D (exact dB values are dependent on the noisy
    instance). On average, our results are $0.34$dB better than BM3D.}
    \label{fig:smallpeppers}
\end{figure}

\subsection{Bayesian bounds} \citet{levin2010natural}
derive bounds on how well any denoising algorithm can perform. The bounds are
dependent on the patch size, where larger patches lead to better results. For
large patches and low noise, tight bounds cannot be estimated.  On the image
depicted in Figure~\ref{fig:smallpeppers}a (a down-sampled and cropped
version of the image ``Peppers'') and for noise level $\sigma=75$, the
theoretically best achievable result using patches of size $12\times12$ is
estimated to be $0.07$dB better than BM3D ($23.86$dB for BM3D and $23.93$ for
the estimated bound). 

We tested an MLP trained on $\sigma=75$ as well as other methods (including
BM3D) on the same image and summarize the results in
Table~\ref{tab:peppersbounds}. We used $100$ different noisy versions of the
same clean image and report the worst, best and average results obtained.  For
BM3D, we obtain results that are in agreement with those obtained
by \citet{levin2010natural}, though we note that the difference between the
worst and best result is quite large: Approximately $1$dB. The high variance in
the results is due to the fact that the test image is relatively small and the
noise variance quite high. The results obtained with BLSGSM and KSVD are also
in agreement with those reported by \citet{levin2010natural}. NLSC achieves
results that are much worse than those obtained by BM3D on this image and this
noise level. EPLL achieves results that are on par with those achieved by BM3D.

BM3D achieves a mean PSNR of $23.51$dB and our MLP achieves a mean PSNR of
$23.85$dB, an improvement of $0.34$dB. Visually, the difference is noticeable,
see Figure~\ref{fig:smallpeppers}.  This is a much greater improvement than was
estimated to be possible by \citet{levin2010natural}, using patches of size
$12\times12$.  This is possible because of the fact that we used larger
patches. \citet{levin2010natural} were unable to estimate tight
bounds for larger patch sizes because of their reduced density in the dataset
of clean patches.

\citet{levin2010natural} BM3D as a method that uses patches of size
$12\times12$. However, BM3D is a two-step procedure. It is true that BM3D uses
patches of size $12\times12$ (for noise levels above $\sigma=40$) in its first
step. However, the second step of the procedure effectively increases the
support size: In the second step, the patches ``see beyond'' what they would
have seen in the first step, but it is difficult to say by how much the support
size is increased by the second step.  Therefore, a fairer comparison would
have been to compare the estimated bounds against only the first step of BM3D.
If only the first step of BM3D is used, the mean result is $22.35$dB.
Therefore, if the constraint on the patch sizes is strictly enforced for BM3D,
the difference between the theoretically best achievable result and BM3D is
larger than suggested by \citet{levin2010natural}.

\subsection{Bayesian bounds with unlimited patch size} More recently,
bounds on denoising quality achievable using any patch size have been 
suggested~\citep{levin2012patch}. This was done by extrapolating bounds similar
to those suggested by \citet{levin2010natural} to larger patch sizes (including
patches of infinite size). For $\sigma=50$ and $\sigma=75$, the bounds lie
$0.7$dB and $1$dB above the results achieved by BM3D, respectively. The
improvements of our approach over BM3D on these noise levels (estimated on
$2500$ images) are $0.32$dB and $0.36$dB, respectively.  Our approach therefore
reaches respectively $46\%$ and $36\%$ of the remaining possible improvement
over BM3D. Furthermore, \citet{levin2012patch} suggest that increasing the
patch size suffers from a law of diminishing returns. This is particularly true
for textured image content: The larger the patch size, the harder it becomes to
find enough training data. \citet{levin2012patch} therefore suggest that
increasing the patch size should be the most useful for smooth image content.
The observation that our method performs much better than BM3D on images with
smooth areas (see middle row in Figure~\ref{fig:imagecomparison}) is in
agreement with this statement. The fact that image denoising is cursed with a
law of diminishing returns also suggests that the remaining available
improvement will be increasingly difficult to achieve. However,
\citet{levin2012patch} suggest that patch-based denoising can be improved
mostly in flat areas and less in textures ones. Our observation that the MLP
performs particularly well in areas with complicated structure (such as on the
bottom image in Figure~\ref{fig:imagecomparison} or both images in
Figure~\ref{fig:mandrill}) shows that large improvements over BM3D on images
with complicated textures are possible.

\section{Results: comparison on non-AWG noise}
\label{sec:othernoise}
Virtually all denoising algorithms assume the noise to be AWG.  However, images
are not always corrupted by AWG noise. Noise is not necessarily additive,
white, Gaussian and signal independent.  For instance in some situations, the
imaging process is corrupted by Poisson noise (such as photon shot noise).
Denoising algorithms which assume AWG noise might be applied to such images
using some image transform \citep{makitalo2011poissonoptimal}. 
Rice-distributed noise, which occurs in magnetic resonance imaging, can be
handled similarly~\citep{foi2011ricenoise}.

\begin{figure*}
  \centering
  \begin{tabular}{ccc}
    \includegraphics[width=0.32\textwidth]{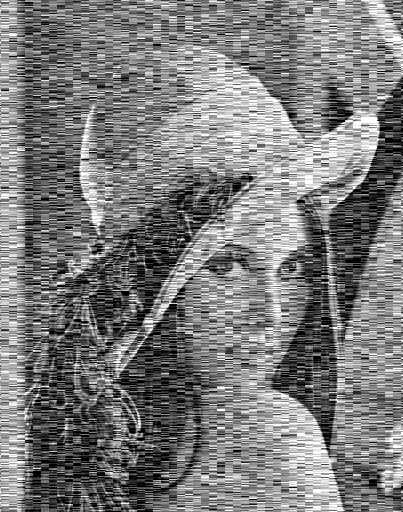} &
    \includegraphics[width=0.32\textwidth]{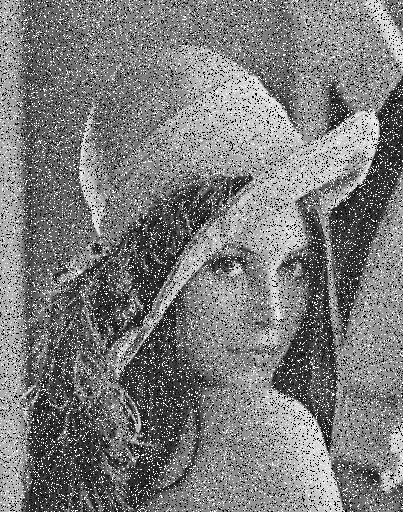} &
    \includegraphics[width=0.32\textwidth]{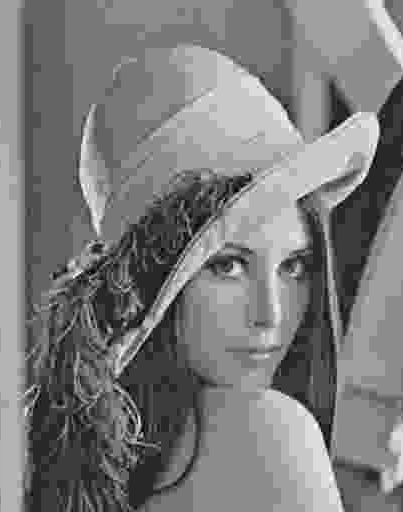} \\
    \small{``stripe'' noise: $14.68$dB} & \small{s $\&$ p noise: $12.41$dB} & \small{JPEG quantization: $27.33$dB}\\
    \includegraphics[width=0.32\textwidth]{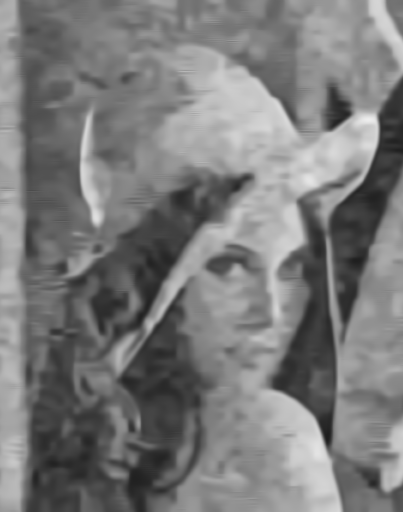} &
    \includegraphics[width=0.32\textwidth]{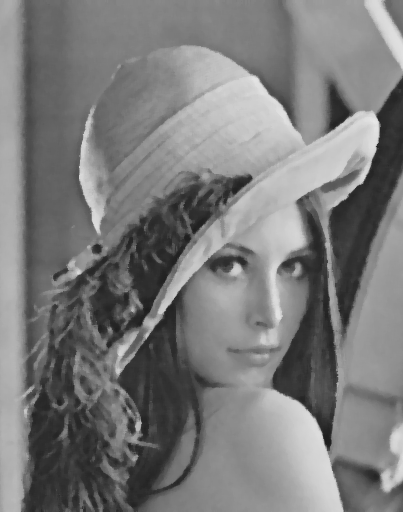} &
    \includegraphics[width=0.32\textwidth]{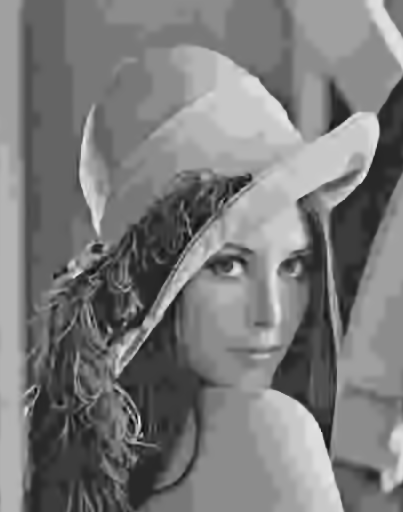} \\
    \small{BM3D~\citep{dabov2007image}: $24.38$dB} & \small{median filtering: $30.33$dB} & \small{SA-DCT~\citep{foi2007pointwise}: $28.96$dB} \\
    \includegraphics[width=0.32\textwidth]{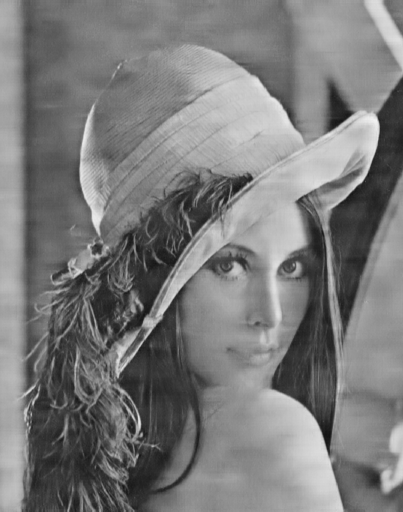} &
    \includegraphics[width=0.32\textwidth]{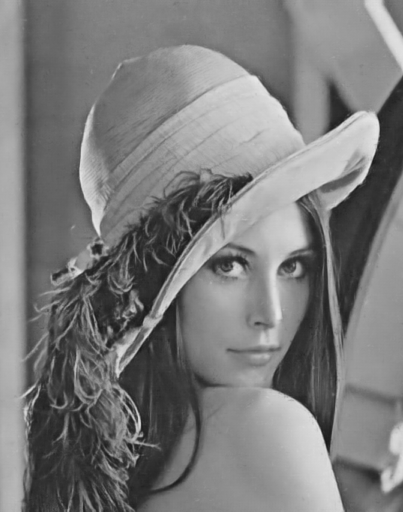} &
    \includegraphics[width=0.32\textwidth]{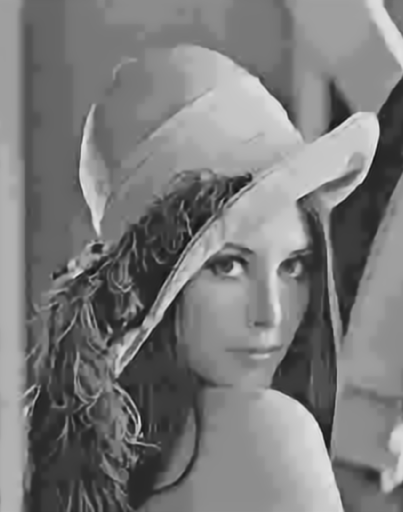} \\
    \small{MLP: $\mathbf{30.11}$dB} & \small{MLP: $\mathbf{35.08}$dB} & \small{MLP:$\mathbf{29.42}$dB}
  \end{tabular}
  \caption{Comparison of our method to other on stripe noise (left),
  salt-and-pepper noise (middel) and JPEG quantization artifacts (right). BM3D
  is not designed for stripe noise.}
  \label{fig:lenaothernoise}
\end{figure*}

In most cases however, it is more difficult or even impossible to find
Gaussianizing transforms. In such cases, a possible solution is to create a
denoising algorithm specifically designed for that noise type. MLPs allow us to
effectively learn a denoising algorithm for a given noise type, provided that
noise can be simulated. In the following, we present results on three noise
types that are different from AWG noise.  We make no effort to adapt our
architecture or procedure in general to the specific noise type but rather use
an architecture that yielded good results for AWG noise (four hidden layers of
size $2047$ and input and output patches of size $17\times17$).  

\medskip 

\subsection{Stripe noise}
It is often assumed that image data contains structure, whereas the noise is 
uncorrelated and therefore unstructured. In cases where the noise also exhibits
structure, this assumption is violated and denoising results become poor. We
here show an example where the noise is additive and Gaussian (with
$\sigma=50$), but where $8$ horizontally adjacent noise values have the same
value.

Since there is no canonical denoising algorithm for this noise, we choose BM3D
as the competitor.  An MLP trained on 82 million training examples outperforms
BM3D for this type of noise, see left column of
Figure~\ref{fig:lenaothernoise}.

\medskip

\subsection{Salt and pepper noise}
When the noise is additive Gaussian, the noisy image value is still correlated
to the original image value. With salt and pepper noise, noisy values are not
correlated with the original image data. Each pixel has a probability $p$ of
being corrupted.  A corrupted pixel has probability $0.5$ of being set to $0$;
otherwise, it is set to highest possible value ($255$ for $8$-bit images). We
show results with $p=0.2$. 

A common algorithm for removing salt and pepper noise is median filtering. We
achieved the best results with a filter size of $5\times 5$ and symmetrically
extended image boundaries.
We also experimented with BM3D (by varying the value of $\sigma$) and achieved
a PSNR of $25.55$dB. An MLP trained on 88 million training examples outperforms
both methods, see middle column of Figure~\ref{fig:lenaothernoise}.  

The problem of removing salt and pepper noise is reminiscent of the in-painting
problem, except that it is not known which pixels are to be in-painted. If one
assumes that the positions of the corrupted pixels are known, the pixel values
of the non-corruped pixels can be copied from the noisy image, since these are
identical to the ground truth values. Using this assumption, we achieve
$36.53$dB with median filtering and $38.64$dB with the MLP.

\medskip

\subsection{JPEG quantization artifacts}
Such artifacts occur due to the JPEG image compression algorithm.  The
quantization process removes information, therefore introducing noise.
Characteristics of JPEG noise are blocky images and loss of edge clarity. This
kind of noise is not random, but rather completely determined by the input
image. In our experiments we use JPEG's quality setting $Q=5$, creating visible
artifacts.

A common method to enhance JPEG-compressed images is to shift the images,
re-apply JPEG compression, shift back and
average~\citep{nosratinia2001enhancement}. This method achieves a PSNR of
$28.42$dB on our image. We also compare against the state-of-the-art in JPEG
de-blocking~\citep{foi2007pointwise}.

An MLP trained on 58 million training examples with that noise outperforms both
methods, see right column of Figure~\ref{fig:lenaothernoise}. In fact, the
method described by \citet{nosratinia2001enhancement} achieves an improvement
of only $1.09$dB over the noisy image, whereas our method achieves an
improvement of $2.09$dB. SA-DCT~\citep{foi2007pointwise} achieves an
improvement of $1.63$dB.

\begin{figure*}
  \centering
  \begin{tabular}{cc}
    \includegraphics[width=0.45\textwidth]{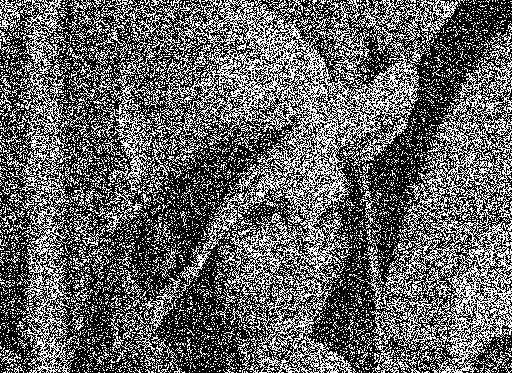} &
    \includegraphics[width=0.45\textwidth]{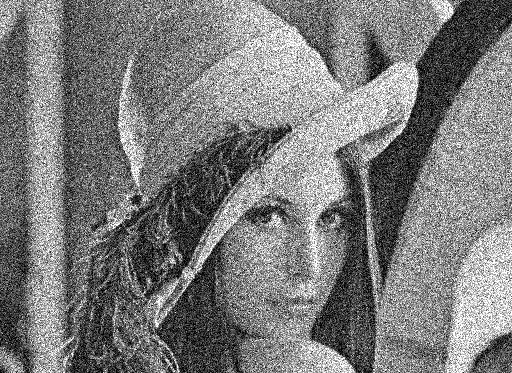} \\
    noisy, peak=1, PSNR: $2.87$dB & noisy, peak=20, PSNR: $14.53$dB \\
    \includegraphics[width=0.45\textwidth]{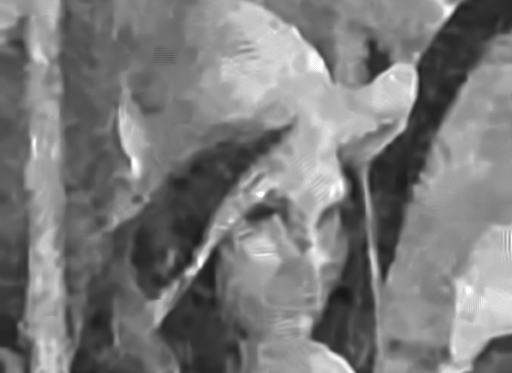} &
    \includegraphics[width=0.45\textwidth]{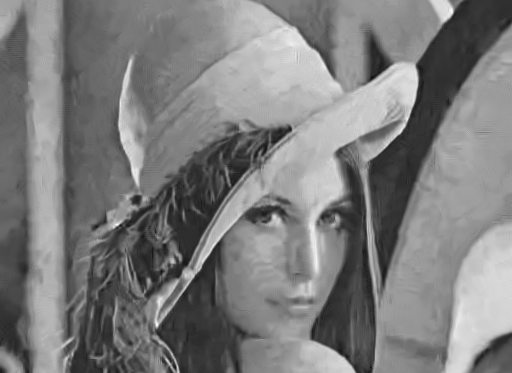} \\
    {\small GAT+BM3D~\citep{makitalo2012optimal}} & {\small GAT+BM3D~\citep{makitalo2012optimal}} \\
    PSNR: $22.90$dB & PSNR: $29.36$dB \\
    \includegraphics[width=0.45\textwidth]{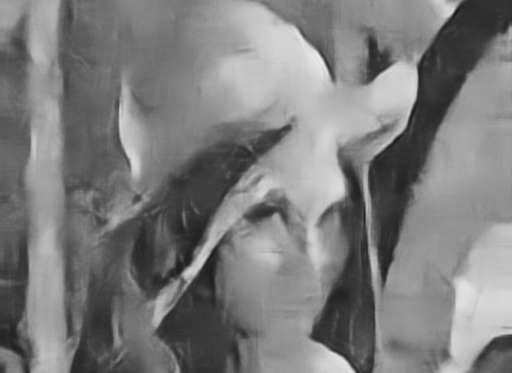} &
    \includegraphics[width=0.45\textwidth]{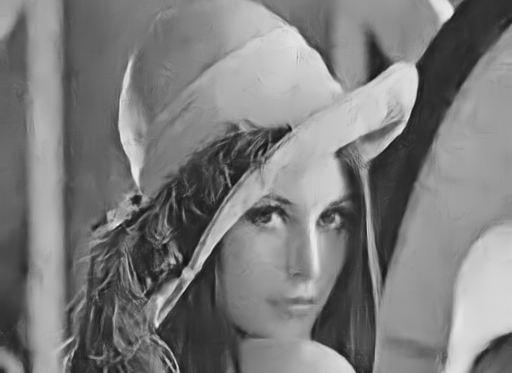} \\
    MLP: $\mathbf{24.26}$dB & MLP: $\mathbf{29.89}$dB
  \end{tabular}
  \caption{Comparison of our method to GAT+BM3D~\citep{makitalo2012optimal} on
  images corrupted with mixed Poisson-Gaussian noise, which occurs in
  photon-limited imaging.}
  \label{fig:mixedpoisson}
\end{figure*}

\medskip
\subsection{Mixed Poisson-Gaussian noise}
In photon-limited imaging, observations are usually corrupted by mixed
Poisson-Gaussian noise~\citep{makitalo2012optimal,luisier2011image}.
Observations are assumed to come from the following model:
\begin{equation}
\label{eq:poisson}
z = \alpha p + n , 
\end{equation}
where $p$ is Poisson-distributed with mean $x$ and $n$ is Gaussian-distributed
with mean $0$ and variance $\sigma^{2}$.  One can regard $x$ to be the
underlying ``true'' image of which one wishes to make a noise-free observation.
To generate a noisy image from a clean one, we follow the setup used by by
\citet{makitalo2012optimal} and \citet{luisier2011image}: We take the clean
image and scale it to a given peak value, giving us $x$.
Applying~(\ref{eq:poisson}) gives us a noisy image $z$.

Two canonical approaches exist for denoising in the photon-limited setting: (i)
Applying a variance stabilizing transform on the noisy image, running a
denoising algorithm designed for AWG noise (such as BM3D) on the result and
finally applying the inverse of the variance stabilizing transform, and (ii)
designing a denoising algorithm specifically for mixed Poisson-Gaussian noise.
GAT+BM3D~\citep{makitalo2012optimal} is an example of the first approach,
whereas UWT/BDCT PURE-LET~\citep{luisier2011image} is an example of the second
approach.  In the case where a variance-stabilizing transform such as the
Anscombe transformation or the generalized Anscombe transformatoin
(GAT)~\citep{starck1998image} is applied, the difficulty lies in the design of
the inverse transform
\citep{makitalo2009inversion,makitalo2011closed,makitalo2011optimal,makitalo2012poisson}.
Designing a denoising algorithm specifically for Poisson-Gaussian noise is also
a difficult task, but can potentially lead to better results.

\begin{table}
  \centering
  \begin{tabular}{lc|cccc}
    image & peak & GAT+BM3D & UWT/BDCT (Foi) & UWT/BDCT (Luisier) & MLP \\
    \hline
    \hline
    Barbara & 1 & 20.83dB & - & 20.79dB & \textbf{21.44}dB \\
    Barbara & 20 & \textbf{27.52}dB & - & 27.33dB & 26.08dB \\
    \hline
    Cameraman & 1 & 20.34dB & 20.35dB & 20.48dB & \textbf{21.66}dB \\
    Cameraman & 20 & 26.83dB & 25.92dB & \textbf{26.93}dB & \textbf{26.93}dB \\
    \hline
    Lena & 1 & 22.96dB & 22.83dB & - & \textbf{24.26}dB \\
    Lena & 20 & 29.39dB & 28.46dB & - & \textbf{29.89}dB \\
    \hline
    Fluo.cells & 1 & 24.54dB & 25.13dB & 25.25dB & \textbf{25.56}dB \\
    Fluo.cells & 20 & 29.66dB & 29.47dB & 31.00dB & 29.98dB \\
    \hline
    Moon & 1 & 22.84dB & - & \textbf{23.49}dB & 23.48dB\\
    Moon & 20 & 25.28dB & - & \textbf{26.33}dB & 25.71dB
  \end{tabular}
  \caption{Comparison of MLPs against two competing methods on mixed
  Poisson-Gaussian noise.  The MLPs perform particularly well when the noise is
  strong (peak = $1$), but are also competitive on lower noise.}
  \label{tab:poisson}
\end{table}
Our approach to denoising photon-limited data is to train an MLP on data
corrupted with mixed Poisson-Gaussian noise.  We trained an MLP on noisy images
using a peak value of $1$ and another MLP for peak value $20$, both on $60$
million examples. For the Gaussian noise, we set $\sigma$ to the peak value
divided by $10$, again following the setup used by \citet{makitalo2012optimal}
and \citet{luisier2011image}. We compare our results against
GAT+BM3D~\citep{makitalo2012optimal}, which is considered state-of-the-art. We
compare on further images in Table~\ref{tab:poisson}. For UWT/BDCT
PURE-LET~\citep{luisier2011image}, we noticed a discrepancy between the results
reported by \citet{makitalo2012optimal} and by \citet{luisier2011image} and
therefore report both.  We see that the MLPs outperform the state-of-the art on
image ``Lena'' in both settings.

%%%%%%%%%%%%%%%%%%%%%%%%%%%%%%%%%%%%%%%%%%%%%%%%%%%%%%%%%%%%%%%%%%%%%%%%%%%%%%%%

\begin{figure}[hbp]
  \centering
  \begin{tabular}{cc}
    \includegraphics[width=0.48\columnwidth]{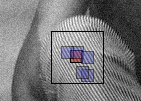} &
    \includegraphics[width=0.48\columnwidth]{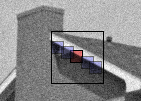} \\
    (a) & (b)
  \end{tabular}
  \caption{Block matching: The goal of the procedure is to find the patches
  most similar to the reddish (``reference'') patch. The neighbors (blueish
  patches) have to be found within a search region (represented by the larger
  black bounding box). Patches can overlap. Here, the procedure was applied on
  (a) the ``Barbara'' image and (b) the ``House'' image, both corrupted with
  AWG noise with $\sigma=10$.}
  \label{fig:bm001}
\end{figure}
\section{Combining BM3D and MLPs: block-matching MLPs}
Many recent denoising algorithms rely on a block-matching procedure. This most
notably includes BM3D \citep{dabov2007image}, but also NLSC
\citep{mairal2010non}. The idea is to find patches similar to a reference patch
and to exploit these ``neighbor'' patches for better denoising.  More
precisely, the procedure exploits the fact that the noise in the different
patches is \emph{independent}, whereas the (clean) image content is
\emph{correlated}. Figure~\ref{fig:bm001} shows the effect of the procedure on
two images.

Since this technique has been used with so much success, we ask the question:
Can MLPs be combined with a block matching procedure to achieve better results?
In particular, can we achieve better results on images where we perform rather
poorly compared to BM3D and NLSC, namely images with repeating structure?  To
answer this question, we train MLPs that take as input not only the
\emph{reference} patch, but also its $k$ nearest \emph{neighbors} in terms of
$\ell_{2}$ distance. We will see that such \emph{block-matching MLPs} can
indeed achieve better on images with repeating structure. However, they also
sometimes achieve worse results than plain MLPs and do not achieve better
results on average.

\subsection{Differences to previous MLPs} Previously, we trained MLPs to
take as input one noisy image patch and to output one denoised image patch. The
best results were achieved when the input patch size was $39\times39$ and the
output patch was of size $17\times17$. Now, we train MLPs to take as input $k$
noisy patches of size $13\times13$ or $17\times17$ and to output one noisy
patch of the same size. The block matching procedure has to be performed for
each training pair, slowing down the training procedure by approximately a
factor of $2$. One could also imagine MLPs taking as input $k$ patches and
providing $k$ patches as output, but we have been less successful with that
approach. In all our experiments, we used $k=14$. The architecture of the MLP
we used had four hidden layers; the first hidden layer was of size $4095$ and
the remaining three were of size $2047$. We discuss the training procedure
in~\citet{burgerjmlr2}.

We note that our block-matching procedure is different from the one employed by
BM3D in a number of ways: (i) We always use the same number of neighbors,
whereas BM3D chooses all patches whose distance to the reference patch is
smaller than a given threshold, up to a maximum of $32$ neighbors, (ii) BM3D is
a two-step approach, where the denoising result of the first step is merely
used to find better neighbors in the second step. We find neighbors directly in
the noisy image. (iii) When the noise level is higher than $\sigma=40$, BM3D
employs ``coarse pre-filtering'' in the first step: patches are first
transformed (using a 2D wavelet or DCT transform) and then hard-thresholded.
This is already a form of denoising and helps to find better neighbors. We
employ no such strategy. (iv) BM3D has a number of hyper-parameters (patch and
stride sizes, type of 2D transform, thresholding and matching coefficients).
The value of the hyper-parameters are different for the two steps of the
procedure. We have fewer hyper-parameters, in part due to the fact that our
procedure consists of a single step. We also choose to set the search stride
size to the canonical choice of $1$.

\subsection{Block-matching MLPs vs.~plain MLPs}
\paragraph{Results on 11 standard test images:} Table~\ref{tab:bmresults}
summarizes the results achieved by an MLP using block matching with $k=14$,
patches of size $13\times13$ and $\sigma=25$. We omit KSVD and EPLL from the
comparison because the block-matching MLP and the plain MLP both outperform the
two algorithms on every image. The mean result achieved on the $11$ images is
$0.07$dB higher for the block-matching MLP than for the plain MLP. The
block-matching MLP outperforms NLSC on $8$ images, whereas the plain MLP
outperforms NLSC on $7$ images. The block-matching MLP outperforms the plain
MLP on $7$ images. The improvement on the plain MLP is the largest on images
Barbara, House and Peppers (approximately $0.25$dB on each). The largest
decrease in performance compared to the plain MLP is observed on image Lena (a
decrease of $0.11$dB). We see that the block-matching procedure is most useful
on images with repeating structure, as found in the images ``Barbara'' and
``House''. However, both BM3D and NLSC achieve results that are far superior 
to the block-matching MLP on image ``Barbara''.

\begin{table} 
  \centering
  \begin{tabular}{l||ccccc}
    %image & BM3D\cite{dabov2007image} & NLSC\cite{mairal2010non} & MLP & BM-MLP\\
    image & BM3D & NLSC & MLP & BM-MLP\\
    \hline\hline 
    Barbara &  \textbf{30.67}dB  &  \textit{30.50}dB  &  29.52dB  &  29.75dB  \\  
    Boat &  29.86dB  &  29.86dB  &  \textbf{29.95}dB  &  \textit{29.92}dB  \\  
    C.man &  29.40dB  &  29.46dB  &  \textit{29.60}dB  &  \textbf{29.67}dB  \\  
    Couple &  29.68dB  &  29.63dB  &  \textbf{29.75}dB  &  \textit{29.73}dB  \\  
    F.print &  \textbf{27.72}dB  &  27.63dB  &  \textit{27.67}dB  &  27.63dB  \\  
    Hill &  29.81dB  &  29.80dB  &  \textit{29.84}dB  &  \textbf{29.87}dB  \\  
    House &  \textit{32.92}dB  &  \textbf{33.08}dB  &  32.52dB  &  32.75dB  \\  
    Lena &  32.04dB  &  31.87dB  &  \textbf{32.28}dB  &  \textit{32.17}dB  \\  
    Man &  29.58dB  &  29.62dB  &  \textit{29.85}dB  &  \textbf{29.86}dB  \\  
    Montage &  \textbf{32.24}dB  &  \textit{32.15}dB  &  31.97dB  &  32.11dB  \\  
    Peppers &  30.18dB  &  \textit{30.27}dB  &  30.27dB  &  \textbf{30.53}dB
  \end{tabular}
  \caption{Block matching MLP compared to plain MLPs and other algorithms for
  $\sigma=25$}
  \label{tab:bmresults}
\end{table}

\begin{figure}[htbp]
  \centering
    \includegraphics[width=\columnwidth]{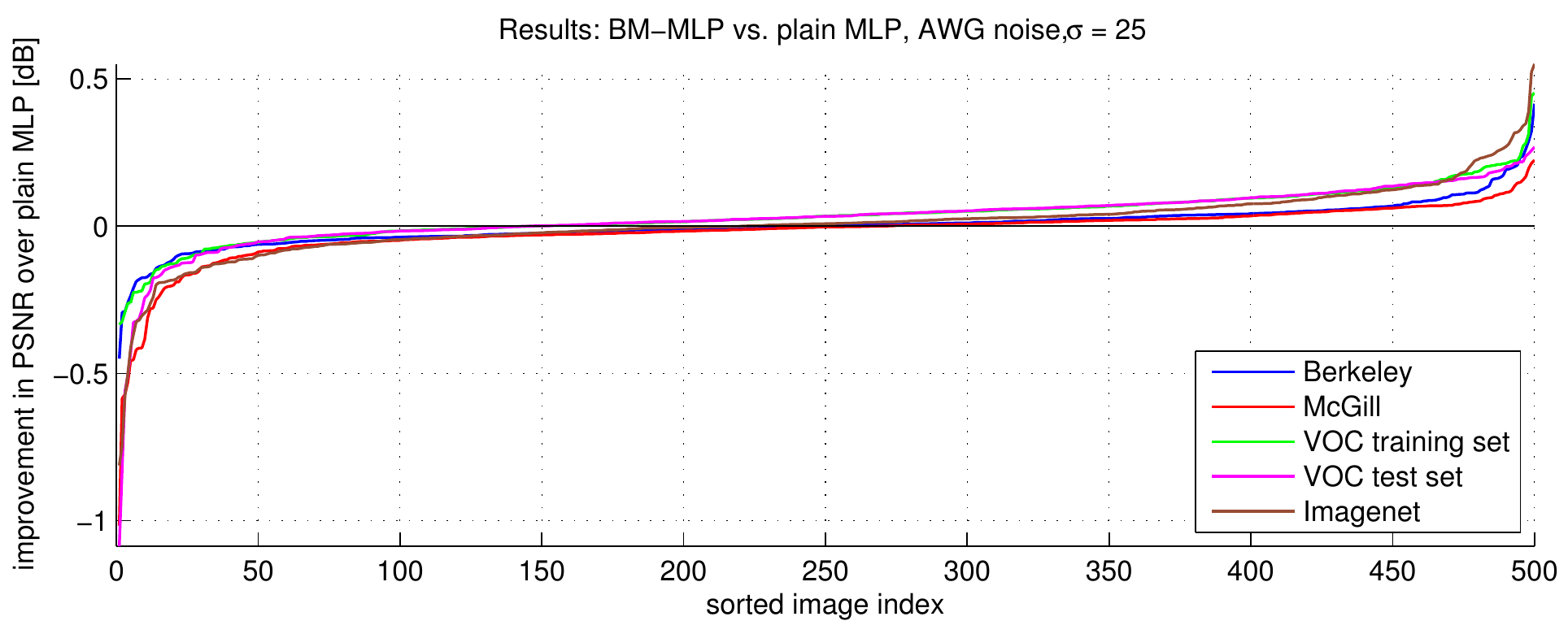}
    \caption{Results of the block-matching MLP compared to the plain MLP on five datasets of 500 images}
  \label{fig:comparison005}
\end{figure}

\paragraph{Results on larger test sets:} The block-matching MLP
outperforms the plain MLP on $1480$ ($59.2\%$) of the $2500$ images, see
Figure~\ref{fig:comparison005}. The average improvement over all datasets is
$0.01$dB. The largest improvement was on the VOC training set ($0.03$dB). On
the McGill dataset, the block-matching MLP was worse by $0.01$dB. The
block-matching MLP and the plain MLP therefore achieve approximately equal
results on average.

\begin{figure}[htbp]
  \centering
    \begin{tabular}{cc}
      \includegraphics[width=0.45\columnwidth]{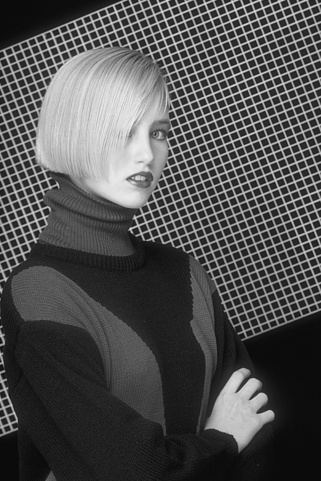} &
      \includegraphics[width=0.45\columnwidth]{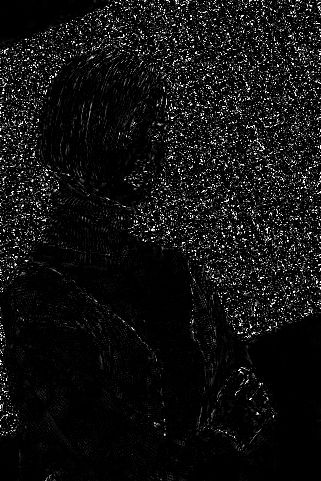} \\
      (a) & (b)
    \end{tabular}
    \caption{The MLP with block-matching outperforms the plain MLP on this
    image. (a) Clean image (b) regions where the block-matching MLP is better
    are highlighted.}
  \label{fig:bmmlpgain}
\end{figure}

On image $198023$ in the Berkeley dataset, the MLP with block-matching
outperforms the plain MLP by $0.42$dB. This is an image similar to the
``Barbara'' images in that it contains a lot of regular structure, see
Figure~\ref{fig:bmmlpgain}.

On image $004513$ in the VOC test set, see Figure~\ref{fig:imagecomparison},
the MLP with block-matching performs $1.09$dB worse than the plain MLP. This
can be explained by the fact that the block-matching MLP uses smaller patches,
making it blind to low frequency noise, resulting in a decrease in
performance on images with smooth surfaces.

\paragraph{Conclusion:} On average, the results achieved with a
block-matching MLP are almost equal to those achieved by a plain MLP. Plain
MLPs perform better on images with smooth surfaces whereas the block-matching
MLPs provide better results on images with repeating structure. However,
combining MLPs with the block-matching procedure did not allow us to outperform
BM3D and NLSC on image ``Barbara''. We emphasize that the block-matching MLPs
use less information as input than the plain MLPs, yet still achieve results
that are comparable on average. Block-matching is a search procedure and
therefore cannot be learned by a feed-forward architecture with few layers.

%%%%%%%%%%%%%%%%%%%%%%%%%%%%%%%%%%%%%%%%%%%%%%%%%%%%%%%%%%%%%%%%%%%%%%%%%%%%%%%%%%
\section{Code}
We make available a \textsc{Matlab} toolbox allowing to denoise images with
our trained MLPs on CPU at \url{http://people.tuebingen.mpg.de/burger/neural_denoising/}.
The script \texttt{demo.m} loads the image ``Lena'', adds AWG noise with
$\sigma=25$ on the image and denoises with an MLP trained on the same noise
level. Running the script produces an output similar to the following
\begin{verbatim}
  >> demo
  Starting to denoise...
  Done! Loading the weights and denoising took 121.4 seconds
  PSNRs: noisy: 20.16dB, denoised: 32.26dB
\end{verbatim}
and display the clean, noisy and denoised images. Denoising an image is
performed using the function \texttt{fdenoiseNeural}:
\begin{verbatim}
  >> im_denoised = fdenoiseNeural(im_noisy, noise_level, model);
\end{verbatim}
The function takes as input a noisy image, the level of noise and a struct
containing the step size and the width of the Gaussian window applied on
denoised patches.
\begin{verbatim}
  >> model = {};
  >> model.step = 3;
  >> model.weightsSig = 2;
\end{verbatim}

%%%%%%%%%%%%%%%%%%%%%%%%%%%%%%%%%%%%%%%%%%%%%%%%%%%%%%%%%%%%%%%%%%%%%%%%%%%%%%%%%%
\section{Discussion and Conclusion}
In this paper, we have described a learning-based approach to image denoising.
We have compared the results achieved by our approach against other algorithms
and against denoising bounds, allowing us to draw a number of conclusions.

\paragraph{Comparison against state-of-the-art algorithms:}
\begin{itemize}
  \item \textbf{KSVD:} We compared our method against
  KSVD~\citep{elad2006image} on $11$ test images and for all noise levels. KSVD
  outperforms our method only on image Barbara with $\sigma=10$.
  \item \textbf{EPLL:} We outperform EPLL~\citep{zoran2011learning} on more than
  $99\%$ of the $2500$ test images on $\sigma=25$, and by $0.35$dB on average.
  For all other noise levels and $11$ test images, we always outperform EPLL.
  \item \textbf{NLSC:} We outperform NLSC~\citep{mairal2010non} more
  approximately $80\%$ of the $2500$ test images on $\sigma=25$, and by
  $0.16$dB on average. The higher the noise level, the more favorably we
  perform against NLSC. NLSC has an advantage over our method on images with
  repeating structure, such as Barbara and House. However, at high noise
  levels, this advantage disappears.
  \item \textbf{BM3D:} We outperform BM3D~\cite{dabov2007image} on approximately
  $92\%$ of the $2500$ test images on $\sigma=25$, and by $0.29$dB on average.
  Otherwise, the same conclusions as for NLSC hold: The higher the noise level,
  the more favorably we perform against BM3D. BM3D has an advantage over our
  method on images with repeating structure, such as Barbara and House.
  However, at high noise levels, this advantage disappears.
\end{itemize}
Our method compares the least favorably compared to other methods on the lowest
noise level ($\sigma=10$), but we still achieve an improvement of $0.1$dB over
BM3D on that noise level.

\paragraph{Comparison against denoising bounds:}
\begin{itemize}
  \item \textbf{Clustering-based bounds} Our results exceed the bounds
  estimated by \citet{chatterjee2010denoising}. This is possible because we
  violate the ``patch cluster'' assumption made by the authors. 
  %We conclude
  %that the patch cluster assumption is not a reasonable assumption to make in
  %order to estimate denoising bounds. 
  In addition,
  \citet{chatterjee2010denoising} suggest that there is almost no room for
  improvement over BM3D on images with complex textures. We have seen that is
  not the case: Our approach is often significantly better than BM3D on images
  with complex textures.
  \item \textbf{Bayesian patch-based bounds} \citet{levin2010natural} estimate
  denoising bounds in a Bayesian setting, for a given patch size.  Our results
  are superior to these bounds This is possible because we use larger patches
  than is assumed by \citet{levin2010natural}. The same authors also suggest that
  image priors should be the most useful for denoising at medium noise levels,
  but not so much at high noise levels. Yet, our method achieves the greatest 
  improvements over other methods at high noise levels.
  
  Similar bounds estimated for patches of infinite size are estimated by
  \citet{levin2012patch}. We make important progress toward reaching these
  bounds: Our approach reaches almost half the theoretically possible gain over
  BM3D. \citet{levin2012patch} agree with \citet{chatterjee2010denoising}
  that there is little room for improvement on patches with complex textures.
  We have seen that this is not the case.
\end{itemize}

\paragraph{Comparison on other noise types:} We have seen that our method can
be adapted to other types of noise by merely switching the training data. We
have shown that we achieve good results are on stripe noise, salt-and-pepper
noise, JPEG quantization artifacts and mixed Poisson-Gaussian noise. In the
latter two cases we seem to be competitive with the state-of-the-art.

\paragraph{Block-matching MLPs:} We have also seen that results can sometimes
be improved a little further using a block-matching procedure. However, this
comes at the cost of a more complicated training procedure and longer training
and test times. In addition, the block-matching procedure is highly
\emph{task-specific}: It has been shown to work well on AWG noise, but it is
not clear that it is useful for all kinds of noise. In addition, plain MLPs
could potentially be used for other low-level vision tasks. It is not clear
that the block-matching procedure is useful for other tasks. We here face an
often encountered conundrum: Is it worth exploiting task-specific knowledge?
This often leads to better results, at the cost of more engineering.

\paragraph{Computation time:} 
Denoising an image using an MLP takes approximately a minute on CPU and less
than $5$ seconds on GPU. This is not as fast as BM3D, but much faster than
approaches that require learning a dictionary, such as KSVD or NLSC which can
take almost an hour per image (on CPU).

\paragraph{Training procedure:} Part 2 of this paper~\citep{burgerjmlr2}, 
describes our training procedure in detail and shows the importance of various
factors influencing the quality of the
results, such as the size of the training corpus, the architecture of the
multi-layer perceptrons and the size of the input and output patches. We show
that some setups lead to surprisingly bad results and provide an explanation
for the phenomena.

\paragraph{Understanding denoising:} Also not discussed in this paper is
the operating principle of the multi-layer perceptrons: How do they achieve
denoising? Trained neural networks are often seen as ``black boxes'', but we
will see in \citet{burgerjmlr2} that in this case, the behavior can be understood,
at least to some extent.

\paragraph{Outlook:} On some images, our method outperforms BM3D by more
than $1.5$dB and NLSC by more than $3$dB, see Section~\ref{sec:results}. Our
method therefore seems to have a clear advantage over other methods on some
images. However, we have seen that our approach sometimes achieves results that
are much worse than the previous state-of-the-art. This happens especially on
images with a lot of regular structure, such as the image ``Barbara''. Our
attempt to ameliorate the situation using a block-matching procedure was only
partially successful. A question therefore begs to be asked: Can we find an
approach that achieves state-of-the-art results on every image?  An approach
combining several algorithms, such as the one proposed by \citet{jancsary2012}
might be able to solve that problem.

\vskip 0.2in
%\bibliography{sample}
%\bibliographystyle{plain}
\bibliography{mybib}

\end{document}